\documentclass{article}
\usepackage[nusbwlab]{iclr2025_conference}
\usepackage{times}
\renewcommand{\bwlheadlab}{}   % header banner shows only "Preprint"

\usepackage{amsmath}
\usepackage{amssymb}
\usepackage{amsfonts}
\usepackage{amsmath,amsfonts,bm}

\def\eqref#1{equation~\ref{#1}}
\def\1{\bm{1}}

\DeclareMathAlphabet{\mathsfit}{\encodingdefault}{\sfdefault}{m}{sl}
\SetMathAlphabet{\mathsfit}{bold}{\encodingdefault}{\sfdefault}{bx}{n}

\usepackage{hyperref}
\usepackage{url}
\usepackage{booktabs}
\usepackage{nicefrac}
\usepackage{microtype}
\usepackage{xcolor}
\usepackage{graphicx}
\graphicspath{{figures/}}
\usepackage{multirow}
\usepackage{xspace}
\usepackage{pifont}
\usepackage{float}
\usepackage{placeins}

\newcommand{\cmark}{\ding{51}}
\newcommand{\xmark}{\ding{55}}
\newcommand{\ours}{\texttt{SinAE}\xspace}

\hypersetup{
  pdftitle={SinAE: High-Fidelity Structural Autoencoding across Molecules, Crystals and Protein Backbones},
  pdfauthor={Yuxuan Ren, Fan Yang, Jianhua Yao, Yatao Bian}
}

\title{SinAE: High-Fidelity Structural\\
Autoencoding across Molecules,\\
Crystals and Protein Backbones}

\author{Yuxuan Ren$^{1}$\quad Fan Yang$^{2}$\quad Jianhua Yao$^{2}$\quad Yatao Bian$^{1,}$\thanks{Corresponding author} \\
$^{1}$Department of Computer Science, National University of Singapore \\
$^{2}$Tencent AI for Life Sciences Lab \\
{\small\texttt{\{yuxuan.ren,ybian\}@nus.edu.sg},\quad
\texttt{\{fionafyang,jianhuayao\}@tencent.com}}}

\begin{document}
\maketitle

\begin{abstract}
Molecules, crystals and protein backbones share a description in terms
of labelled sites in three-dimensional space, yet their structural
models largely use separate representations and architectures.
A common latent interface must preserve fine geometry while remaining
useful for generation. We introduce \ours, a vanilla-Transformer
autoencoder with typed-site tokens and an iterative flow-matching
decoder. A common reconstruction objective supports molecules,
periodic structures and protein C$\alpha$ backbones without explicit
equivariant or graph operators. The design achieves sub-milliangstrom
molecular reconstruction, milliangstrom-level periodic reconstruction
and approximately $0.01$\,\AA{} protein C$\alpha$ RMSD. A matched
molecular comparison links flow decoding to a roughly hundred-fold
reduction in reconstruction error, and physical round trips assess
preservation beyond coordinate accuracy. Joint molecule--crystal
training improves reconstruction and generation in both domains, and three-domain sharing reveals a fidelity trade-off. Standard Diffusion
Transformer priors over the learned latents support competitive
molecular, crystal and protein-backbone generation. These results
establish a reusable autoencoding design that connects high-fidelity
structural representation with latent generative modelling across
domains.
\end{abstract}

\section{Introduction}
\label{sec:intro}

Learning to represent and generate three-dimensional atomic structures
supports drug discovery, materials design and protein engineering.
Molecules, crystals and protein backbones share a simple description:
categorical site labels and positions in three-dimensional space.
Their modelling pipelines nevertheless use different geometric
representations and architectures. Molecular models commonly use
message passing or equivariant operations~\citep{gilmer2017neural,schutt2018schnet,thomas2018tensor};
crystal models account for periodic boundaries and lattice-basis
invariance~\citep{xie_crystal_2022,jiao_crystal_nodate}; and protein
models use residue frames or pair
representations~\citep{yim_frameflow_2023,watson_rfdiffusion_2023,jumper2021alphafold}.
A common structural interface would make architectures reusable across
these settings and allow heterogeneous datasets to contribute to shared
parameters. This is particularly relevant when individual datasets,
such as QM9 and MP-20, contain only about $130$\,k molecules and
$45$\,k crystals, respectively~\citep{ramakrishnan2014quantum,xie_crystal_2022}.

\begin{figure}[t]
    \centering
    \includegraphics[width=\linewidth]{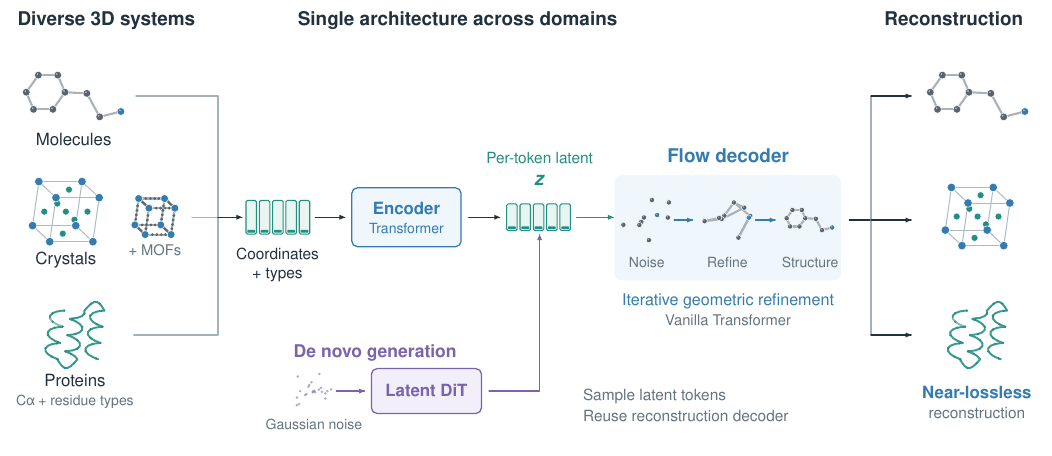}
    \caption{\textbf{\ours overview.} A common Transformer
    architecture represents molecules, crystals/MOFs, and protein
    C$\alpha$ backbones with per-token latents. The iterative flow
    decoder reconstructs structures from encoded latents and is reused
    to decode samples from a separately trained latent DiT. The token interface supports joint molecule--crystal training.}
    \label{fig:teaser}
\end{figure}

Two routes address this fragmentation. Direct generators learn
structures in coordinate space or through spatial
tokenisation~\citep{morehead_zatom1_2026,lu2026unified,zhang2025unigenxunifiedgenerativefoundation}.
They must model composition and geometry together with the different
structural requirements of each domain. Latent approaches separate
structural reconstruction from distribution learning: an autoencoder
encodes an input structure and a generative prior models its
codes~\citep{xu2023geoldm,xie_crystal_2022,joshi_all-atom_nodate}.
This separation makes reconstruction fidelity central. Errors introduced
by the autoencoder can alter local geometry and the physical quantities
associated with a structure before the prior is trained.

Existing autoencoders establish complementary capabilities.
UAE-3D demonstrates near-lossless molecular
reconstruction~\citep{luo_towards_2025}, whereas ProteinAE uses
iterative decoding for protein backbones~\citep{li_proteinae_2025}.
ADiT provides a shared vanilla-Transformer autoencoder and latent
Diffusion Transformer (DiT) for molecules and
crystals~\citep{joshi_all-atom_nodate}. These advances raise a concrete
question: can a common autoencoding design retain fine geometric
detail across molecules, periodic structures and protein backbones,
while providing latents that a generic prior can use for generation?
Answering it requires both accurate reconstruction and evidence that
the latent interface remains useful for sampling.

We introduce \ours, a vanilla-Transformer autoencoder that combines a
common typed-site interface with iterative flow-matching decoding
(Table~\ref{tab:capabilities}). Molecules and periodic structures use
atom types and coordinates; proteins use residue types and
C$\alpha$ coordinates. Three lattice-vector tokens extend the interface
to crystals and metal--organic frameworks (MOFs). Following
 diffusion-based decoding~\citep{chen_diffusion_2025}, the decoder
refines noisy structures conditioned on per-token latents. Coordinate
reconstruction, categorical type prediction and light KL regularisation
use the same objective weights across domains, without geometric
auxiliary losses specific to an individual domain. Architecture reuse
is tested across all settings, and parameter sharing is tested
explicitly through joint training.

\begin{table}[t]
    \centering
    \small
    \setlength{\tabcolsep}{3pt}
    \caption{\textbf{Capability comparison of \ours\ with latent autoencoder baselines.}
    \cmark: supported; \xmark: not supported.
    \emph{Vanilla Transformer}: backbone uses neither equivariant
    operators nor graph / message-passing.
    \emph{Joint type+coord.}: model generates discrete atom-type or
    sequence labels jointly with continuous coordinates.
    \emph{Sub-m\AA{} recon}: reconstruction RMSD below
    $10^{-3}$\,\AA{} on a molecular benchmark.}
    \label{tab:capabilities}
    \begin{tabular}{l ccc c c c}
        \toprule
        & \multicolumn{3}{c}{Domains} & Vanilla & Joint & Sub-m\AA{} \\
        \cmidrule(lr){2-4}
        Method & Mol & Crystal & Protein & Transformer & type+coord. & recon \\
        \midrule
        GeoLDM~\citep{xu2023geoldm}             & \cmark & \xmark & \xmark & \xmark & \cmark & \xmark \\
        CDVAE~\citep{xie_crystal_2022}          & \xmark & \cmark & \xmark & \xmark & \cmark & \xmark \\
        ADiT~\citep{joshi_all-atom_nodate}      & \cmark & \cmark & \xmark & \cmark & \cmark & \xmark \\
        ProteinAE~\citep{li_proteinae_2025}     & \xmark & \xmark & \cmark & \cmark & \xmark & \xmark \\
        \midrule
        \textbf{\ours\ (ours)} & \cmark & \cmark & \cmark & \cmark & \cmark & \cmark \\
        \bottomrule
    \end{tabular}
\end{table}

We evaluate this design through an evidence chain that connects
reconstruction to generative use. Cross-domain benchmarks establish
geometric fidelity, a matched decoder comparison tests the contribution
of iterative flow decoding, and physical round trips and timing
profiles assess preservation and computational cost. Joint-training
experiments then examine the benefits and limits of shared parameters.
Finally, separately trained DiT priors~\citep{peebles_dit_2023} use the
frozen autoencoder's latents for molecular, crystal and protein-backbone
generation. Together, these experiments test whether a reusable
structural interface can support both precise reconstruction and
competitive generation.

\section{Results}
\label{sec:experiments}
\label{sec:analysis}

\subsection{A Common Autoencoding Interface}
\label{subsec:overview}
\label{subsec:featurization}

We represent each structure by coordinates
$\mathbf{X}\in\mathbb{R}^{N\times3}$ and categorical site labels
$\mathbf{A}\in\{1,\ldots,K\}^N$. A site is an atom for molecules
and crystals, or a residue at its C$\alpha$ position for proteins.
Periodic systems append the three lattice vectors as dedicated tokens,
giving $N'=N+3$; otherwise $N'=N$. Each token sums a linear coordinate
projection and a learned type embedding at width $d_{\mathrm{model}}$.
Aperiodic inputs are centred; random SO(3) rotations and aperiodic
translation augmentation encourage orientation robustness. No canonical frames, pairwise invariant
features, equivariant layers, or graph message passing are used.

\begin{figure}[t]
    \centering
    \includegraphics[width=\linewidth]{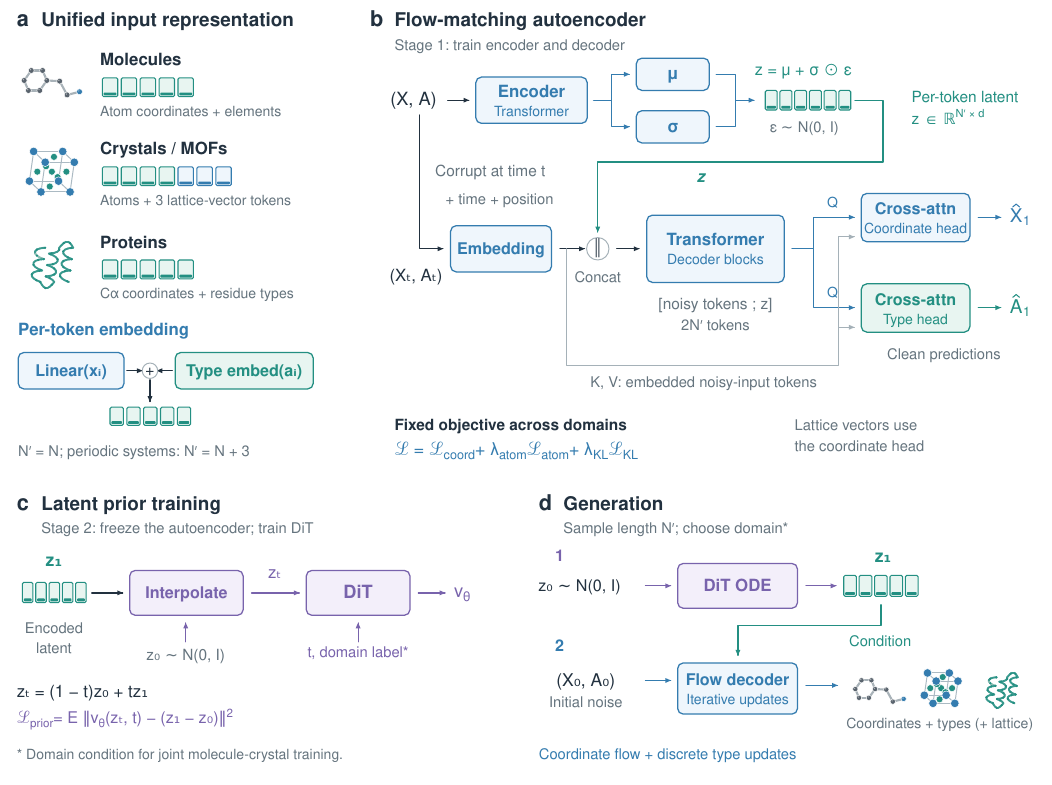}
    \caption{\textbf{\ours architecture and procedures.}
    \textbf{(a)} Unified inputs for molecules, periodic crystals/MOFs,
    and protein C$\alpha$ backbones; periodic systems add three lattice
    tokens. \textbf{(b)} Stage 1 trains the stochastic encoder and
    flow-matching decoder with coordinate, type, and KL terms.
    \textbf{(c)} Stage 2 freezes the autoencoder and trains a latent
    DiT prior. \textbf{(d)} Generation samples latent tokens and
    iteratively decodes a structure. Domain conditioning is used for
    joint molecule--crystal training.}
    \label{fig:main}
\end{figure}

For a structure $\mathcal G=(\mathbf X,\mathbf A)$, the encoder
defines $q_\phi(\mathbf z\mid\mathcal G)$ and the decoder reconstructs
$\mathcal G$ conditioned on $\mathbf z$. Training has two sequential
stages (Figure~\ref{fig:main}). First, the encoder and conditional flow
decoder are trained jointly for reconstruction. The autoencoder is then
frozen and a latent DiT prior is trained on encoder outputs. The same
decoder is used for reconstruction and for decoding samples from the
prior.

We assess geometric fidelity on QM9 and GEOM-Drugs molecules, MP-20
crystals, QMOF structures and held-out CASP14/CASP15 protein backbones.
QM9 and MP-20 have single-domain and jointly trained configurations;
GEOM-Drugs, QMOF and the main protein benchmarks use separate models.
A three-domain autoencoder tests broader parameter sharing for
reconstruction. Dataset and training details are given in Methods (Section~\ref{sec:method}).
\subsection{Reconstruction Fidelity}
\label{subsec:exp_recon}

Tables~\ref{tab:recon_nonprot} and~\ref{tab:recon_prot} report
reconstruction accuracy. The joint molecule--crystal model reaches
$0.0002$\,\AA{} on QM9 and $0.0017$\,\AA{} on MP-20, with
$100\%$ structure match on both datasets. The separately trained
GEOM-Drugs model reaches $0.0006$\,\AA{} and $100\%$ match.
QMOF extends the periodic evaluation to larger hybrid structures,
with $0.0021$\,\AA{} RMSD and $99.9\%$ match. These are
sub-milliangstrom errors for the molecular benchmarks and
milliangstrom-level errors for the periodic benchmarks.

\noindent\textbf{Comparison with existing autoencoders.}
On QM9, \ours matches UAE-3D's $0.0002$\,\AA{} RMSD; on
GEOM-Drugs it reaches $0.0006$\,\AA{} versus $0.0008$\,\AA{}.
Structure match improves over ADiT from $94.6\%$ to $100\%$ on
QM9 and from $84.5\%$ to $100\%$ on MP-20.
Section~\ref{subsec:decoder_control} tests the flow decoder in a
controlled comparison. On CASP14/CASP15, \ours achieves
$0.010$--$0.014$\,\AA{} protein-backbone CA-RMSD
(Table~\ref{tab:recon_prot}).
\begin{table}[t]
    \centering
    \footnotesize
    \setlength{\tabcolsep}{3pt}
    \caption{\textbf{Reconstruction on molecules and crystals.}
    \textbf{(a)} structure match rate and Kabsch-aligned coordinate
    RMSD on molecule benchmarks. \textbf{(b)} coordinate RMSD and
    pymatgen structure match rate on periodic benchmarks (MP-20; QMOF
    as an additional secondary domain). ``\ours (joint)'' on MP-20
    denotes the model trained jointly on QM9 + MP-20. Periodic RMSD
    is also in \AA{}.}
    \label{tab:recon_nonprot}
    \begin{minipage}[t]{0.52\linewidth}
    \centering
    \textbf{(a) Molecules.}\\[2pt]
    \begin{tabular}{llcc}
    \toprule
    Dataset & Method & Match\,$\uparrow$ & RMSD\,(\AA)\,$\downarrow$ \\
    \midrule
    \multirow{5}{*}{QM9}
    & GeoLDM & 87.2  & 0.1830 \\
    & ADiT   & 94.6  & 0.089 \\
    & UAE-3D & 100.0 & 0.0002 \\
    & \textbf{\ours (QM9)}  & \textbf{100.0} & 0.0007 \\
    & \textbf{\ours (joint)}& \textbf{100.0} & \textbf{0.0002} \\
    \midrule
    \multirow{3}{*}{\shortstack[l]{GEOM-\\Drugs}}
    & GeoLDM & 82.3  & 0.2526 \\
    & UAE-3D & 100.0 & 0.0008 \\
    & \textbf{\ours (GEOM)} & \textbf{100.0} & \textbf{0.0006} \\
    \bottomrule
    \end{tabular}
    \end{minipage}\hfill
    \begin{minipage}[t]{0.46\linewidth}
    \centering
    \textbf{(b) Periodic systems.}\\[2pt]
    \begin{tabular}{llcc}
    \toprule
    Dataset & Method & RMSD\,$\downarrow$ & Match\,$\uparrow$ \\
    \midrule
    \multirow{3}{*}{MP-20}
    & ADiT & 0.1827 & 84.50 \\
    & \textbf{\ours (MP20)}  & 0.0041 & \textbf{100.0} \\
    & \textbf{\ours (joint)} & \textbf{0.0017} & \textbf{100.0} \\
    \midrule
    QMOF & \textbf{\ours (QMOF)} & \textbf{0.0021} & \textbf{99.9} \\
    \bottomrule
    \end{tabular}
    \end{minipage}
\end{table}
\begin{table}[t]
    \centering
    \footnotesize
    \setlength{\tabcolsep}{3pt}
    \caption{\textbf{Protein C$\alpha$ reconstruction RMSD ($\downarrow$, \AA).}
Mean over evaluation samples; \ours values are bold. \ours uses
chains truncated to $256$ residues; baseline C$\alpha$ values from
Table~1 of \citet{li_proteinae_2025} retain their source evaluation
protocols. Full mean$\,\pm\,$std.\ values are in
Appendix~\ref{app:recon_prot_full}.}
    \label{tab:recon_prot}
    \begin{tabular}{lccccc}
        \toprule
        & \multicolumn{3}{c}{CASP14} & \multicolumn{2}{c}{CASP15} \\
        \cmidrule(lr){2-4} \cmidrule(lr){5-6}
        Method & T & T-dom & oligo & TS-dom & oligo \\
        \midrule
        CHEAP & 11.15 & 8.99 & 9.93 & 10.22 & 9.22 \\
        ESM3 VQ-VAE & 1.02 & 0.66 & 3.08 & 1.23 & 1.94 \\
        DPLM-2 & 1.99 & 1.87 & 2.70 & 3.31 & 3.50 \\
        ProteinAE & 0.23 & 0.22 & 0.31 & 0.28 & 0.37 \\
        \midrule
        \textbf{\ours} & \textbf{0.010} & \textbf{0.011} & \textbf{0.013} & \textbf{0.013} & \textbf{0.014} \\
        \bottomrule
    \end{tabular}
\end{table}
\subsection{Controlled Decoder Comparison}
\label{subsec:decoder_control}

The QM9 ablation holds the encoder, latent representation, decoder
depth and width, and training schedule and step budget fixed. The
regression variant directly predicts $(\mathbf X_1,\mathbf A_1)$;
the flow variant conditions on noisy inputs and time and reconstructs
iteratively. Flow decoding reduces RMSD from $0.072$ to
$0.0007$\,\AA{}, an approximately $103\times$ improvement.
This comparison tests the iterative, noise-conditioned decoder as a
design choice, including its training objective and inference procedure.

The result links the flow-decoding design to a large improvement in
reconstruction accuracy while retaining the same encoder and latent
interface. The following diagnostics examine what this improvement
means for physical preservation and refinement cost.

\subsection{Physical Distortion in Encoding--Decoding Round Trips}
\label{subsec:roundtrip}

Table~\ref{tab:roundtrip} measures how encoding and decoding affect
physical quantities. On QM9, \ours preserves mean UFF strain energy
at $2.70$\,kcal/mol and retains molecular stability of $0.9633$
from a reference value of $0.9667$; ADiT reaches $0.5733$.
For the $100$ MP-20 structures, reference and reconstructed hull
energies both round to $0.031\pm0.047$\,eV/atom. Relaxation RMSD
changes from $0.064$ to $0.075$\,\AA{}, compared with
$0.362$\,\AA{} for ADiT, while the stable count changes from
$18$ to $15$. These results show that \ours preserves molecular
stability and crystal energy statistics through the encoding--decoding
round trip.

\begin{table}[t]
\centering
\small
\setlength{\tabcolsep}{7pt}
\caption{\textbf{Round-trip physical diagnostics.}
Top: QM9, mean UFF strain energy in kcal/mol and stability fractions.
Bottom: a diagnostic subset of $100$ MP-20 training structures;
hull energies in eV/atom, stable entries are counts, and relaxation
RMSD is in \AA{}. Validity is in \%.}
\label{tab:roundtrip}
\begin{tabular}{lccc}
\toprule
QM9 & Reference & \ours & ADiT \\
\midrule
Mean strain energy & 2.70 & 2.70 & 8.31 \\
Atom stability & 0.9928 & 0.9923 & 0.9183 \\
Molecule stability & 0.9667 & 0.9633 & 0.5733 \\
\midrule
MP-20 & Reference & \ours & ADiT \\
\midrule
Validity & 99.0 & 99.0 & 95.0 \\
$E_{\mathrm{hull}}$ (mean$\pm$std.) & $0.031\pm0.047$ & $0.031\pm0.047$ & $0.30\pm0.40$ \\
Stable count & 18 & 15 & 0 \\
Relaxation RMSD & 0.064 & 0.075 & 0.362 \\
\bottomrule
\end{tabular}
\end{table}

\subsection{Refinement Steps and Inference Cost}
\label{subsec:efficiency}

With all models timed on the same single GPU, the QM9 component profile takes $3456.22$\,ms for
latent sampling and $25.65$\,ms for ten-step decoding per call;
decoding contributes $0.74\%$ of the combined time. ADiT takes
$3431.76$\,ms and $3.62$\,ms with one decoder pass. End-to-end
QM9 throughput is $0.0553$\,s/sample for \ours, compared with
$0.0536$ for ADiT, $0.0353$ for SemlaFlow, and $0.1220$ for
EQGAT-diff. Appendix~\ref{app:profiling} gives the full profiles and
their configurations. Figure~\ref{fig:nfe} in that appendix relates
reconstruction precision to the decoder step count.

Reconstruction RMSD falls sharply between one and four decoder
evaluations and saturates near six on both QM9 and MP-20. The QM9
sweep reaches approximately $0.0003$\,\AA{} and MP-20 reaches
$0.0016$\,\AA{}. Iterative refinement therefore concentrates its
largest gains in the early steps. The joint molecular--crystal
generation and component-profiling configurations are specified in
Appendix~\ref{app:profiling}.

\begin{figure}[t]
    \centering
    \includegraphics[width=0.74\linewidth]{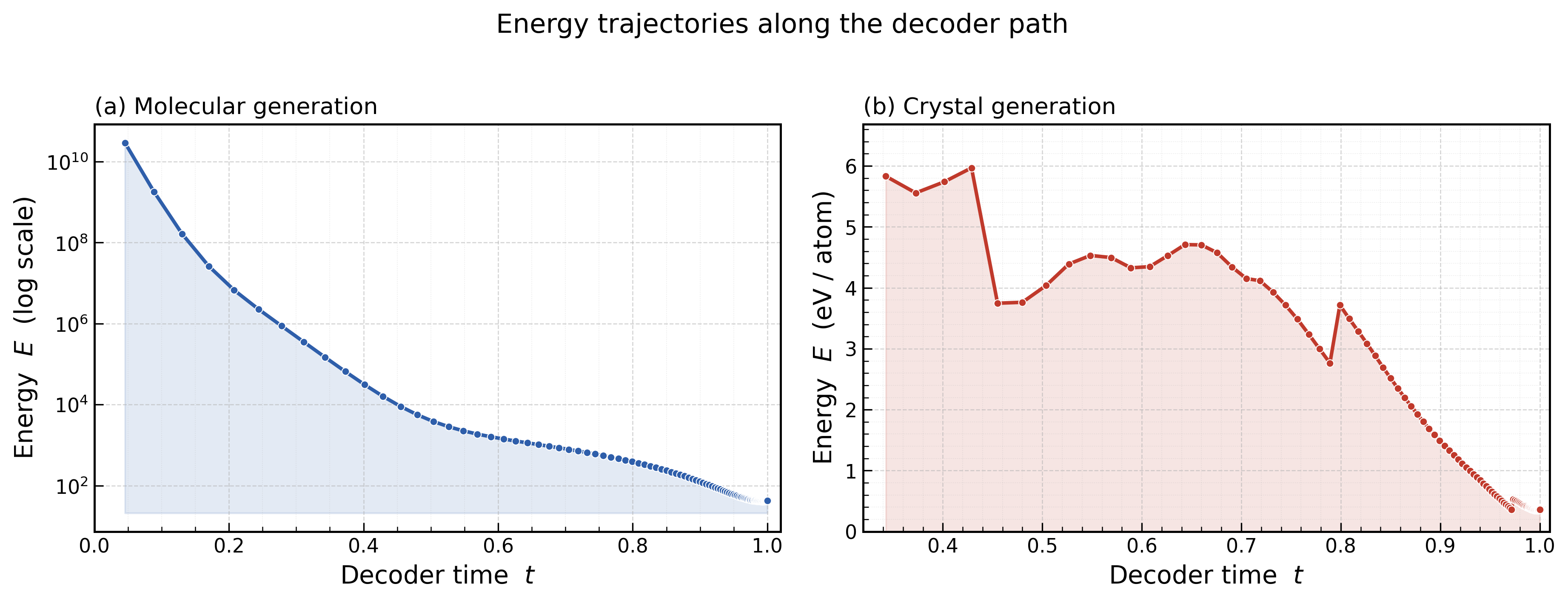}\par
    \vspace{2pt}
    \includegraphics[width=0.74\linewidth]{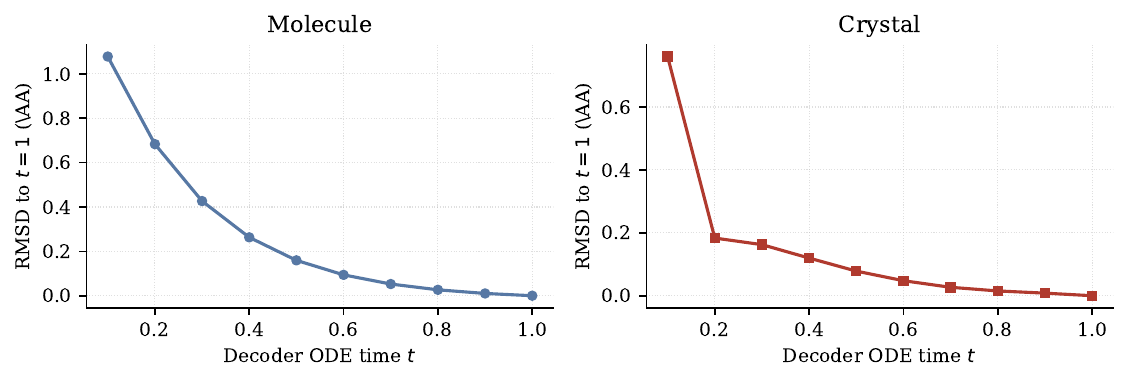}
    \caption{\textbf{Decoder trajectories for prior-sampled latents.}
    Top: molecular force-field and crystal CHGNet energies. Bottom:
    Kabsch-aligned RMSD to the final decoded state. Energy continues to
    change after most coordinate movement has occurred.}
    \label{fig:trajectory}
\end{figure}

\paragraph{How structures evolve during decoding.}
\label{subsec:energy}
Figure~\ref{fig:trajectory} follows generation from DiT-sampled
latents. Molecular force-field energies and crystal CHGNet energies
decrease along the displayed decoder paths. RMSD to the final decoded
state becomes small around $t\approx0.8$, while energy continues to
change near the endpoint. This is consistent with small late-stage
coordinate refinements affecting energy more than RMSD. Energy is not
used as a training target.

\subsection{Benefits and Limits of Domain Sharing}
\label{subsec:sharing}

Joint QM9 + MP-20 training lowers RMSD from $0.0007$ to
$0.0002$\,\AA{} on QM9 and from $0.0041$ to $0.0017$\,\AA{} on
MP-20 (Table~\ref{tab:recon_nonprot}). Both populations benefit
from a shared autoencoder. Section~\ref{sec:generation} examines the
corresponding gains in sampling from their shared latent prior.

\paragraph{Three-domain shared parameters.}
A single three-domain autoencoder reconstructs molecules, crystals,
and proteins, with crystal RMSD $0.0057$\,\AA{} and protein
CA-RMSD $0.0116$\,\AA{}. Adding proteins increases molecule and
crystal errors relative to the two-domain model, while protein error
remains close to the approximately $0.012$\,\AA{} single-domain
result. Extending parameter sharing to three domains thus trades some
molecular and crystal precision for broader coverage.

\paragraph{Qualitative diagnostics.}
Latent projections and within-domain interpolations are given in
Appendices~\ref{app:tsne} and~\ref{app:interpolation}. Molecule and
crystal latents occupy organised regions, and interpolation produces
gradual changes in decoded structures. These views complement the
quantitative reconstruction and joint-training comparisons.

\subsection{Generation from the Learned Representation}
\label{sec:generation}

The reconstruction study establishes the fidelity and sharing
properties of the autoencoder. We now fit standard DiT priors to its
latents and evaluate the quality of decoded samples. This tests the
second role of the representation: an interface on which a generic
generative model can learn molecular, crystal, and protein
distributions. Each benchmark uses the autoencoder--prior configuration
specified in Section~\ref{subsec:datasets}.
For QM9, GEOM-Drugs, and MP-20, we evaluate each \ours configuration
on $10{,}000$ generated samples per dataset. On QM9 and MP-20, this
includes both the single-domain and joint models.

\subsubsection{Molecular Generation}
\label{subsec:gen_mol}

Tables~\ref{tab:gen_mol_qm9} and~\ref{tab:gen_mol_geom} report
QM9 and GEOM-Drugs generation.
The joint \ours model reaches PoseBusters validity $0.97$ on QM9,
compared with $0.94$ for EQGAT-diff, SemlaFlow, and ADiT. The separately trained GEOM-Drugs model reaches $0.93$, compared with $0.88$ for SemlaFlow
and $0.86$ for ADiT. This combines strong geometric quality with
competitive chemical validity, novelty, and diversity. For the shared
QM9 + MP-20 model, PB Valid rises from the QM9-only value of $0.94$
to $0.97$, extending the reconstruction benefit of joint training to
generation. Individual PoseBusters checks are in
Appendix~\ref{app:gen_mol_pb_full}.
\begin{table}[t]
\centering
\begin{minipage}[t]{0.48\linewidth}
    \centering
    \footnotesize
    \setlength{\tabcolsep}{1.5pt}
    \caption{\textbf{Molecule generation on QM9 (Benchmark~I).} ADiT (joint) and \ours~(joint) are
    trained on QM9+MP-20; the remaining models are QM9-only.}
    \label{tab:gen_mol_qm9}
    \begin{tabular}{lcccc}
        \toprule
        Method & Valid\,$\uparrow$ & Unique\,$\uparrow$ & Div\,$\uparrow$ & PB Valid\,$\uparrow$ \\
        \midrule
        EQGAT-diff & 0.99 & \textbf{1.00} & 0.89 & 0.94 \\
        SemlaFlow  & 0.99 & 0.95 & 0.89 & 0.94 \\
        FlowMol    & 0.97 & 0.97 & \textbf{0.92} & 0.92 \\
        ADiT (joint) & 0.97 & 0.97 & 0.90 & 0.94 \\
        \midrule
        \textbf{\ours (QM9)}    & 0.97 & 0.97 & 0.91 & 0.94 \\
        \textbf{\ours (joint)}  & \textbf{0.99} & 0.97 & 0.91 & \textbf{0.97} \\
        \bottomrule
    \end{tabular}
\end{minipage}\hfill
\begin{minipage}[t]{0.48\linewidth}
    \centering
    \footnotesize
    \setlength{\tabcolsep}{1.5pt}
    \caption{\textbf{Molecule generation on GEOM-Drugs
    (Benchmark~II).} \ours~(GEOM) is trained on GEOM-Drugs alone; QM9
    and GEOM-Drugs are never jointly trained. Baseline reevaluations:
    \citet{vonessen_tabasco_2025}, arXiv v1, Table~1.}
    \label{tab:gen_mol_geom}
    \begin{tabular}{lcccc}
        \toprule
        Method & Valid\,$\uparrow$ & Nov\,$\uparrow$ & PB Valid\,$\uparrow$ & Div\,$\uparrow$ \\
        \midrule
        EQGAT-diff & 0.94 & 0.94 & 0.84 & 0.90 \\
        SemlaFlow  & 0.93 & 0.93 & 0.88 & 0.91 \\
        FlowMol    & 0.81 & 0.81 & 0.64 & 0.91 \\
        ADiT (joint) & 0.98 & 0.97 & 0.86 & 0.91 \\
        \midrule
        \textbf{\ours (GEOM)} & \textbf{0.98} & \textbf{1.00} & \textbf{0.93} & \textbf{0.91} \\
        \bottomrule
    \end{tabular}
\end{minipage}
\end{table}
\begin{figure}[t]
    \centering
    \includegraphics[width=1.0\linewidth]{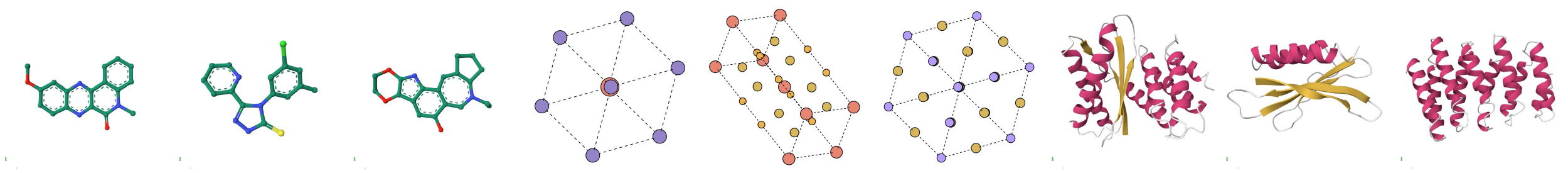}
    \caption{\textbf{Generated samples from \ours.} Unconditional
    samples from the latent DiT prior decoded with the corresponding
    flow-matching decoder (left: small molecules;
    middle: crystals; right: protein backbones). Higher-resolution per-domain visualisations are in
    Appendix~\ref{app:samples_full}.}
    \label{fig:samples}
\end{figure}
\subsubsection{Crystal Generation}
\label{subsec:gen_mat}

Crystal generation targets valid structures that remain stable after
relaxation. Table~\ref{tab:gen_mat} reports MP-20 performance using
S.U.N.\ (Stable, Unique, Novel) and M.S.U.N., with stability thresholds
$E_{\mathrm{hull}}<0$ and $E_{\mathrm{hull}}<0.1$\,eV/atom,
respectively. The joint \ours model reaches S.U.N.\ $5.4\%$ and
M.S.U.N.\ $30.2\%$, compared with $5.3\%$ and $28.2\%$ for
jointly trained ADiT, the closest molecule--crystal latent baseline.
ADiT has higher structural and compositional validity. \ours also
exceeds FlowMM ($2.8\%$ / $22.5\%$) and FlowLLM
($4.7\%$ / $26.3\%$) on S.U.N.\ and M.S.U.N.
Joint training raises these yields from $4.6\%$ and $24.3\%$
for the MP-20-only model, alongside improvements in structural and
compositional validity. Appendix~\ref{app:gen_mat_full} gives the full
validity breakdown and additional crystal-specific baselines. SGFM
achieves S.U.N.\ $6.9\%$ and M.S.U.N.\ $30.3\%$, placing
\ours near the leading M.S.U.N.\ result while using the common
autoencoding design.

\begin{table}[t]
    \centering
    \small
    \setlength{\tabcolsep}{5pt}
    \caption{\textbf{Material generation on MP-20.}
Structure and composition validity, stable rate, M.S.U.N., and S.U.N.
(all in \%, higher is better). ``--'' denotes unavailable values.
Baseline results follow Table~1 of \citet{joshi_all-atom_nodate}.
Full results and additional baselines are in Appendix~\ref{app:gen_mat_full}.}
    \label{tab:gen_mat}
    \begin{tabular}{lccccc}
        \toprule
        Model & Struct\,$\uparrow$ & Comp\,$\uparrow$ & Stable\,$\uparrow$ & M.S.U.N.\,$\uparrow$ & S.U.N.\,$\uparrow$ \\
        \midrule
        CDVAE$^\dagger$  & 100.00 & 86.70 & 1.6  & --    & --   \\
        DiffCSP$^\dagger$  & 100.00 & 83.25 & 5.0  & --    & 3.3  \\
        UniMat$^\dagger$   & 97.2   & 89.4  & --   & --    & --   \\
        FlowMM         & 96.85  & 83.19 & 4.6  & 22.5  & 2.8  \\
        FlowLLM        & 99.94  & 90.84 & 13.9 & 26.3  & 4.7  \\
        Jointly-trained ADiT & 99.74 & 92.14 & 15.4 & 28.2 & 5.3 \\
        \midrule
        \textbf{\ours (MP20-only)} & 98.29 & 88.89 & 14.1 & 24.3 & 4.6 \\
        \textbf{\ours (joint)}     & 99.29 & 90.49 & \textbf{15.8} & \textbf{30.2} & \textbf{5.4} \\
        \bottomrule
        \multicolumn{6}{l}{\scriptsize $^\dagger$\,CDVAE~\citep{xie_crystal_2022},
        DiffCSP~\citep{jiao_crystal_nodate},
        UniMat~\citep{yang_unimat_2024}.}
    \end{tabular}
\end{table}
\subsubsection{Protein-Backbone Generation and Co-Design}
\label{subsec:gen_prot}

The protein autoencoder uses the same architecture as the
molecule--crystal model. Table~\ref{tab:gen_prot} evaluates
unconditional backbone generation through designability, diversity,
DPT, and novelty after inverse folding and re-folding
(Appendix~\ref{app:metrics_protein}). At sampling temperature $0.25$, \ours reaches
$0.96$ designability and $278$ diversity clusters, improving on
ProteinAE's $0.93$ and $204$. At temperature $0.35$, it reaches
$291$ clusters with $0.88$ designability, providing a tunable
quality--diversity trade-off.

\noindent\textbf{Sequence--structure co-design.}
Table~\ref{tab:performance} evaluates jointly generated sequence--backbone
pairs at $100$ and $200$ residues. We fold each sequence emitted by
\ours with ESMFold and compare its predicted backbone with the
co-generated backbone using scTM and scRMSD. \ours reaches scRMSD $3.93$\,\AA{}
at length $100$ and $4.44$\,\AA{} at length $200$, with scTM
$0.83$ and $0.78$, respectively. At length $200$, its scRMSD is
lower than MultiFlow's $4.98$\,\AA{} and ProteinGenerator's
$6.24$\,\AA{}.

\begin{table}[t]
\centering
\begin{minipage}[t]{0.48\linewidth}
    \centering
    \footnotesize
    \setlength{\tabcolsep}{1.5pt}
    \caption{\textbf{Backbone designability.} Des
    (best-of-8 scRMSD$\leq$2.0\,\AA), diversity, DPT, and novelty
    after ProteinMPNN sequence design and ESMFold re-folding.
    Baselines: \citet{li_proteinae_2025}, Table~2;
    ProteinAE uses $\gamma=0.35$.}
    \label{tab:gen_prot}
    \begin{tabular}{lcccc}
        \toprule
        Method & Des\,$\uparrow$ & Div\,$\uparrow$ & DPT\,$\downarrow$ & Nov\,$\downarrow$ \\
        \midrule
        RFdiffusion  & 0.96 & 247 & 0.43 & 0.71 \\
        FrameFlow    & 0.91 & 278 & 0.48 & 0.65 \\
        \midrule
        ESM3         & 0.61 & 127 & 0.37 & 0.84 \\
        DPLM2        & 0.63 & 130 & 0.37 & 0.72 \\
        \midrule
        LSD          & 0.69 & 203 & 0.46 & 0.74 \\
        \midrule
        LatentDiff   & 0.17 & 34  & 0.51 & 0.73 \\
        ProteinAE    & 0.93 & 204 & 0.36 & 0.70 \\
        \textbf{Ours(0.35)} & 0.88 & \textbf{291} & 0.35 & 0.72 \\
        \textbf{Ours(0.25)} & \textbf{0.96} & 278 & 0.36 & 0.74 \\
        \bottomrule
    \end{tabular}
\end{minipage}\hfill
\begin{minipage}[t]{0.50\linewidth}
    \centering
    \footnotesize
    \setlength{\tabcolsep}{1.3pt}
    \caption{\textbf{Sequence--structure co-design} at lengths 100 and 200.
    scTM / scRMSD assess the co-generated pairs.
    $^{*}$: without distillation. Baseline results:
    \citet{chen_apm_2025}, Table~2.}
    \label{tab:performance}
    \begin{tabular}{lcc@{\hskip 2pt}cc}
        \toprule
        \multirow{2}{*}{Method}
         & \multicolumn{2}{c}{Length 100}
         & \multicolumn{2}{c}{Length 200} \\
         \cmidrule(lr){2-3} \cmidrule(lr){4-5}
         & scTM\,$\uparrow$ & scRMSD\,$\downarrow$ & scTM\,$\uparrow$ & scRMSD\,$\downarrow$ \\
        \midrule
        NativePDBs       & 0.91 & 2.98  & 0.88 & 3.24  \\
        \midrule
        ESM3(1.4B)       & 0.72 & 13.80 & 0.63 & 21.18 \\
        MultiFlow$^{*}$  & 0.86 & 4.73  & 0.86 & 4.98  \\
        ProtGen.         & 0.91 & 3.75  & 0.88 & 6.24  \\
        ProtPardelle     & 0.56 & 12.90 & 0.64 & 13.67 \\
        APM              & 0.96 & 1.80  & 0.89 & 4.25  \\
        \midrule
        \textbf{Ours}    & 0.83 & 3.93  & 0.78 & 4.44  \\
        \bottomrule
    \end{tabular}
\end{minipage}
\end{table}

\paragraph{Generative use of the representation.}
Across the benchmarks, a generic DiT prior can use the structural
latents to generate plausible samples in each domain. Molecule--crystal
sharing improves both reconstruction and generation, while the
protein-only model demonstrates reuse of the architecture at residue
scale. The combination gives the autoencoder a practical role as a
common structural interface for reconstruction and sampling.

\FloatBarrier
\section{Discussion}
\label{sec:conclusion}

\ours shows that a common Transformer autoencoding design can preserve
fine structural geometry across molecules, periodic systems and
protein C$\alpha$ backbones, while providing an interface for latent
generation. The shared design combines typed-site tokens, a stochastic
encoder and an iterative flow-matching decoder. Its contribution is a
reusable structural autoencoder supported by reconstruction, physical
preservation, computational profiling and generative evaluation.

The matched QM9 comparison connects the flow-decoding design to a
large reconstruction gain while retaining the encoder, latent interface
and decoder size. Relative to high-fidelity molecular
 autoencoding~\citep{luo_towards_2025} and iterative protein
 autoencoding~\citep{li_proteinae_2025}, \ours extends a common design
across these settings and periodic structures. It also extends the
molecule--crystal latent approach established by
ADiT~\citep{joshi_all-atom_nodate}. The comparison tests the combined
noise-conditioned training and iterative inference design; it does not
isolate the number of inference steps from the training objective.
Round-trip diagnostics further connect coordinate precision to
preservation of molecular stability and crystal energy statistics in
the evaluated samples. The remaining change in crystal stable count
shows why geometric and physical fidelity should be assessed together.

Architecture reuse and shared learning have different empirical
support. The same autoencoding formulation applies to all evaluated
domains, whereas the strongest joint-training results come from QM9
and MP-20. Sharing parameters improves both reconstruction and
generation for these datasets. Adding proteins extends reconstruction
coverage but reduces molecular and crystal precision, indicating that
broader sharing introduces a capacity or optimisation trade-off.
Compute-matched comparisons would be needed to distinguish the effects
of additional data exposure from those of domain composition.

The latent-prior experiments establish that the representation also
supports sampling with a standard DiT. Reconstruction precision and
sample quality are separate properties: a faithful decoder preserves
encoded inputs, while a prior must still learn which latent structures
to sample. Molecular geometric validity, relaxed crystal stability and
protein designability therefore provide complementary tests. The
results support competitive generation across the evaluated settings;
they also retain domain-specific trade-offs, including crystal validity
relative to ADiT and protein sequence--structure consistency.

The current representation retains one latent token per site, with
widths $d=8$ for molecules/crystals and $d=16$ for proteins. Its utility
is demonstrated as a structural interface; stronger information
compression remains untested. Protein reconstruction is restricted to
C$\alpha$ backbones and chains truncated to $256$ residues, with
baseline values retaining their source evaluation protocols. All-atom
proteins and substantially longer chains require further evaluation.
The demonstrated fidelity of a non-equivariant Transformer applies to
the tested training and augmentation settings. Within this scope,
\ours provides a practical foundation for reusing autoencoding and
latent-generation machinery across atomic and residue-scale structures.

\section{Methods}
\label{sec:method}

\subsection{Datasets, Training, and Evaluation Protocol}
\label{subsec:datasets}

\noindent\textbf{Datasets.}
We train and evaluate on standard benchmarks:
\textbf{QM9}~\citep{ramakrishnan2014quantum} and
\textbf{GEOM-Drugs}~\citep{axelrod_geom_2022} for small molecules,
\textbf{MP-20}~\citep{xie_crystal_2022} (from the Materials
Project~\citep{jain_commentary_2013}) for crystals, and
\textbf{AFDB-FS}~\citep{li_proteinae_2025} (a filtered single-chain
subset of the AlphaFold Protein Structure Database) for proteins,
with CASP14/CASP15 held out for protein reconstruction evaluation. As
an additional periodic benchmark we use
\textbf{QMOF}~\citep{rosen_machine_2021} for MOF reconstruction and
generation; the full QMOF results and a head-to-head MOF generation
comparison with ADiT are reported in
Appendix~\ref{app:mof_gen_full}. All splits follow prior
work~\citep{hoogeboom_equivariant_2022,vignac_midi_2023,irwin_semlaflow_2025,xie_crystal_2022,joshi_all-atom_nodate,vankempen2024foldseek};
full specification in Appendix~\ref{app:datasets_full}.

\noindent\textbf{Training configurations.}
The same backbone design is used with domain-appropriate model sizes.
QM9 and MP-20 are evaluated both separately and jointly; GEOM-Drugs,
QMOF, and proteins use separate models for the main benchmarks. A
three-domain autoencoder is evaluated for reconstruction only
(Section~\ref{subsec:sharing}). Autoencoders are trained first, then
frozen before training their latent priors. The joint molecule--crystal model is
trained on a single GPU for approximately $48$ GPU-hours. Hyperparameters and metrics are
in Appendices~\ref{app:implementation_full} and~\ref{app:eval_metrics}.

\subsection{Flow-Matching Autoencoder}
\label{subsec:architecture}
\label{subsec:training}

The pre-norm Transformer encoder produces per-token posterior parameters
$\boldsymbol\mu_i,\boldsymbol\sigma_i$ and samples
$\mathbf z_i=\boldsymbol\mu_i+\boldsymbol\sigma_i\odot\boldsymbol\epsilon_i$,
with $\boldsymbol\epsilon_i\sim\mathcal N(\mathbf0,\mathbf I)$.
The latent $\mathbf z\in\mathbb R^{N'\times d}$ retains site-level
information with compact channels ($d=8$ for the molecule--crystal
model; $d=16$ for proteins).

The decoder receives noisy coordinates and types, with time and
positional embeddings, concatenated with the latent-token sequence.
Shared Transformer blocks process the $2N'$ tokens. Two cross-attention
output heads predict clean coordinates and categorical types using
embedded noisy-input tokens as keys and values, following the
output-head design of Tabasco~\citep{vonessen_tabasco_2025}.
The same coordinate head predicts lattice vectors. Both autoencoders
use four encoder layers and sixteen decoder layers with hidden width
$64$ and eight attention heads. The latent prior is sized separately
for each training configuration (Appendix~\ref{app:impl_arch}). Full
encoder, decoder, and embedding details are in
Appendix~\ref{app:encoder_decoder_details}.

Motivated by diffusion-based decoding~\citep{chen_diffusion_2025}, we
use continuous coordinate flow matching and discrete type flow
matching, with the common objective
\begin{equation}
\mathcal L=\mathcal L_{\mathrm{coord}}
+\lambda_{\mathrm{atom}}\mathcal L_{\mathrm{atom}}
+\lambda_{\mathrm{KL}}\mathcal L_{\mathrm{KL}}.
\label{eq:loss_total}
\end{equation}
The fixed weights are $\lambda_{\mathrm{atom}}=0.1$ and
$\lambda_{\mathrm{KL}}=10^{-6}$. Lattice tokens use the coordinate
objective, so the loss formulation remains the same for periodic and
aperiodic structures.

\paragraph{Coordinate flow matching~\citep{lipman2023flowmatching}.}
With the interpolant
$\mathbf{X}_t = t\,\mathbf{X}_1 + (1-t)\,\boldsymbol{\epsilon}$,
$\boldsymbol{\epsilon}\sim\mathcal{N}(\mathbf{0},\mathbf{I})$,
$t\sim\mathcal{U}(0,1)$, the coordinate head is trained with
$x_1$-prediction:
\begin{equation}
    \mathcal{L}_{\text{coord}} = \mathbb{E}_{\boldsymbol{\epsilon},\,t}
    \!\left[\tfrac{1}{N}\bigl\|\hat{\mathbf{X}}_1^\theta(\mathbf{X}_t,
    \mathbf{A}_t, t, \mathbf{z}) -
    \mathbf{X}_1\bigr\|_2^2\right].
    \label{eq:loss_coord}
\end{equation}

\paragraph{Atom-type discrete flow
matching~\citep{campbell2024generative}.}
Ground-truth atom types are corrupted by interpolating with a uniform
categorical prior,
$\mathbf{A}_t \sim \mathrm{Cat}(t\,\delta(\mathbf{A}_1) +
(1-t)\,\tfrac{1}{K}\mathbf{1})$, and the head is trained with endpoint
cross-entropy:
\begin{equation}
    \mathcal{L}_{\text{atom}} = \mathbb{E}_{t}
    \!\left[-\tfrac{1}{N}\sum_{i=1}^{N} \mathbf{A}_{1,i}^\top \log
    \hat{\mathbf{A}}_{1,i}^\theta(\mathbf{X}_t, \mathbf{A}_t, t,
    \mathbf{z})\right].
    \label{eq:loss_atom}
\end{equation}

The KL term regularises conditional posteriors toward a standard
Gaussian; its closed form is given in
Appendix~\ref{app:training_objective_full}. This combination trains
geometric reconstruction and type prediction through the shared
backbone. The prior subsequently learns the distribution of the encoded
data.

\subsection{Latent Flow Matching for Generation}
\label{subsec:latent_fm}

A vanilla DiT~\citep{peebles_dit_2023} learns the encoded-data
distribution from a Gaussian base. With a frozen encoder sample
$\mathbf z_1$, Gaussian $\mathbf z_0$, and
$\mathbf z_t=(1-t)\mathbf z_0+t\mathbf z_1$, its objective is
\begin{equation}
\mathcal L_{\mathrm{FM}}=
\mathbb E\!\left\|v_\theta(\mathbf z_t,t)
-(\mathbf z_1-\mathbf z_0)\right\|^2.
\end{equation}
Variable-length sequences are padded and masked within each batch.
At inference, lengths follow the dataset-specific size distribution
(Appendix~\ref{app:datasets_full}). The joint molecule--crystal
prior uses domain labels and classifier-free guidance to select the
target domain; other models use the corresponding domain-specific
prior. The joint model uses $100$ Euler steps to sample $\mathbf z_1$
and $10$ flow-matching decoder steps to generate coordinates and types
(Appendix~\ref{app:impl_eval}).
Crystal stability evaluation uses the relaxation protocol in
Appendix~\ref{app:metrics_crystal}.

\subsection{Baseline Selection and Comparison Scope}
\paragraph{Latent structural models.}
GeoLDM~\citep{xu2023geoldm} learns equivariant molecular latents,
while UAE-3D~\citep{luo_towards_2025} demonstrates near-lossless
molecular reconstruction using a Relational Transformer and rotation
augmentation. For periodic structures, CDVAE~\citep{xie_crystal_2022}
combines a latent representation with a diffusion decoder.
ADiT~\citep{joshi_all-atom_nodate} is the closest cross-domain
baseline: it uses vanilla Transformers, a shared molecule--crystal
latent, and a DiT prior.
ProteinAE~\citep{li_proteinae_2025} establishes diffusion autoencoding
for protein backbones. Motivated also by diffusion-based visual
decoding~\citep{chen_diffusion_2025}, \ours studies a common
flow-matching autoencoding design across these structural settings.
Our study connects reconstruction accuracy to physical round-trip
fidelity, decoder cost, parameter sharing, and downstream generation.
The controlled comparison in Section~\ref{subsec:decoder_control}
tests iterative flow decoding within a matched autoencoder setup.

\paragraph{Direct structural generation.}
EDM~\citep{hoogeboom_equivariant_2022},
EQGAT-diff~\citep{le_navigating_2023},
SemlaFlow~\citep{irwin_semlaflow_2025}, and
FlowMol~\citep{dunn_mixed_2024} generate molecules directly in
coordinate space. FlowMM~\citep{miller_flowmm_2024} and
FlowLLM~\citep{sriram_flowllm_2024} address periodic crystal
generation, while FrameFlow~\citep{yim_frameflow_2023} and
RFdiffusion~\citep{watson_rfdiffusion_2023} generate protein backbones
using residue-frame geometry. Cross-domain approaches such as
Zatom-1~\citep{morehead_zatom1_2026}, autoregressive cross-scale
modelling~\citep{lu2026unified}, and
UniGenX~\citep{zhang2025unigenxunifiedgenerativefoundation} explore
alternative unified interfaces. These methods provide important
comparisons for sample quality, whereas our reconstruction experiments
assess preservation of a supplied input structure. Together, these
evaluations measure representation fidelity and downstream sample quality.

\section*{Data and Code Availability}
This study uses the public QM9, GEOM-Drugs, MP-20, QMOF and
AlphaFold Protein Structure Database resources, with CASP14/CASP15
structures for protein reconstruction evaluation. Dataset sources,
processing and splits are described in
Appendix~\ref{app:datasets_full}. The protein reference sets and
assessment procedures are specified in
Appendix~\ref{app:metrics_protein}.
Code is available at
\url{https://github.com/BlueWhaleLab/SinAE}.

\section*{AI-Assisted Manuscript Preparation}
Generative AI tools assisted with manuscript organisation and revision,
interpretation and presentation of existing results, reference lookup,
LaTeX formatting, and figure preparation. These editorial revisions
did not run experiments or generate new experimental results. The
authors remain responsible for the final content.

\bibliography{references}
\bibliographystyle{iclr2025_conference}

\clearpage
\appendix
\section{Broader Impacts}
\label{app:broader_impacts}

\ours is an architectural contribution towards 3D structural
representation learning; its immediate downstream use cases are
molecular, materials, and protein design in academic and industrial
R\&D. Positive societal impacts include accelerating the discovery of
catalysts for carbon capture and clean energy, stable metal--organic
frameworks for gas storage, and therapeutic lead compounds. We note,
however, that the same generative capabilities could in principle be
misused to accelerate the design of toxic small molecules, narcotics,
or proteins with harmful biological activity. The models released
alongside this work are trained exclusively on publicly available
non-hazardous datasets (QM9, GEOM-Drugs, MP-20, QMOF, AFDB-FS), none
of which is enriched for hazardous compounds, and we do not anticipate
\ours conferring a meaningful advantage over existing, widely available
generative tools for adversarial design. We encourage downstream users
to combine generative models of this class with structure-aware
toxicity and bioactivity filters before experimental validation, and we
are not releasing any fine-tuned weights targeted at restricted
chemical or biological endpoints.

\section{Evaluation Metrics}
\label{app:eval_metrics}

This appendix defines every metric reported in the main text and
specifies the tool, formula, or threshold used for each.

\subsection{Reconstruction Metrics}
\label{app:metrics_recon}

\paragraph{Coordinate RMSD (\AA, $\downarrow$).}
For a structure with $N$ atoms, ground-truth coordinates
$\mathbf{X}\in\mathbb{R}^{N\times 3}$, and reconstructed coordinates
$\hat{\mathbf{X}}\in\mathbb{R}^{N\times 3}$, we first align
$\hat{\mathbf{X}}$ to $\mathbf{X}$ using the Kabsch algorithm (optimal
rotation $R^{*}$, translation $t^{*}$ via SVD), then report
\[
\mathrm{RMSD}(\mathbf{X}, \hat{\mathbf{X}}) \;=\;
\sqrt{\tfrac{1}{N}\sum_{i=1}^{N}\bigl\|\mathbf{x}_i -
(R^{*}\hat{\mathbf{x}}_i + t^{*})\bigr\|_2^{2}}.
\]
For proteins we apply this to the $C_\alpha$ trace (CA-RMSD). For
periodic systems (crystals, MOFs), coordinates are first unwrapped into
a single fractional cell and Kabsch alignment is applied in the Cartesian
space.

\paragraph{Structure Match Rate --- Reconstruction (\%, $\uparrow$).}
For molecule \emph{reconstruction} on QM9, the structure match rate
uses \texttt{pymatgen.analysis.molecule\_matcher.\allowbreak
MoleculeMatcher.\allowbreak get\_rmsd} with InChI-based atom mapping
at tolerance $0.01$\,\AA{}. A reconstructed molecule is counted as a
match iff this routine returns a finite RMSD, i.e.\ the predicted 3D
conformation is chemically isomorphic to the reference under the
OpenBabel bond-perception convention (identical atom counts, atomic
numbers, and inferred bond multiset).

For MP-20 crystal reconstruction, we compute the structure match rate
using \texttt{StructureMatcher} from \texttt{pymatgen}, with thresholds
$l_{\mathrm{tol}}=0.3$, $s_{\mathrm{tol}}=0.5$, and
$\mathrm{angle}_{\mathrm{tol}}=10^\circ$. A 100\% match rate means that
every reconstructed crystal is matched to its reference under these
tolerances.

\paragraph{Valid (\%, $\uparrow$).}
Fraction of generated molecules accepted by RDKit's sanitization
pipeline (valid valences, no kekulisation failure, parseable SMILES).

\paragraph{Unique (\%, $\uparrow$).}
Fraction of valid molecules whose canonical SMILES is distinct from the
other molecules in the sampled batch. Computed on the same $N$ samples
used for \textbf{Valid}.

\paragraph{Novelty (\%, $\uparrow$).}
Fraction of valid molecules whose canonical SMILES does not appear in
the training set.

\paragraph{Diversity ($\uparrow$).}
Mean pairwise Tanimoto \emph{dissimilarity} of Morgan fingerprints
(radius 2, 2048 bits) across the valid generated molecules:
\[
\mathrm{Diversity} = 1 - \tfrac{2}{N(N-1)}\sum_{i<j}
\mathrm{Tanimoto}(\mathbf{f}_i, \mathbf{f}_j).
\]

\paragraph{PB Valid (\%, $\uparrow$).}
PoseBusters-Validity: a molecule is PB-valid only if it simultaneously
passes all PoseBusters sanity checks: chemistry, intra-molecular
clashes, bond lengths and angles within empirical distributions. PB
Valid is the most physically restrictive metric in
Tables~\ref{tab:gen_mol_qm9}--\ref{tab:gen_mol_geom} and is the metric
we emphasise for geometric quality.

\subsection{Crystal Generation Metrics}
\label{app:metrics_crystal}

\paragraph{Structure / Composition / Overall Validity (\%, $\uparrow$).}
\emph{Structure validity} counts generated crystals that pass
pymatgen's structure-level sanity (no atomic overlap, finite lattice).
\emph{Composition validity} counts crystals whose composition is
charge-balanced with standard oxidation states. \emph{Overall validity}
is the fraction of samples passing both checks,
\[
V_{\mathrm{overall}}=\frac{100}{M}\sum_{i=1}^{M}
\mathbf{1}\{S_i\}\mathbf{1}\{C_i\},
\]
where $S_i$ and $C_i$ denote the two checks for sample $i$.
We follow the protocol of CDVAE.

\paragraph{Stable and Metastable Rate (\%, $\uparrow$).}
We follow the two-stage relaxation protocol of
ADiT~\citep{joshi_all-atom_nodate}: each generated crystal is first
pre-relaxed with the CHGNet universal machine-learning
potential~\citep{deng_chgnet_2023}, and the pre-relaxed geometry is then
DFT-relaxed. Stability is evaluated against the Matbench-Discovery
convex hull using the final \emph{DFT} energy-above-hull
$E_{\mathrm{hull}}$:
\[
\text{stable}: E_{\mathrm{hull}} < 0.0\text{ eV/atom and }
\#\text{unique elements} \ge 2,
\]
\[
\text{metastable}: E_{\mathrm{hull}} < 0.1\text{ eV/atom and }
\#\text{unique elements} \ge 2.
\]
The $\ge 2$ unique-element constraint matches ADiT and excludes trivial
elemental crystals from stability counting.

\paragraph{S.U.N.\ and M.S.U.N.\ (\%, $\uparrow$).}
\emph{S.U.N.} (Stable, Unique, Novel) is the fraction of sampled
crystals that are simultaneously (i) stable under the above definition,
(ii) structurally unique within the sample batch, and (iii) novel with
respect to the MP-20 training set. Uniqueness and novelty both use
\texttt{pymatgen.StructureMatcher} with the thresholds listed
in Appendix~\ref{app:metrics_recon} for an all-to-all equivalence check.
\emph{M.S.U.N.} replaces ``stable'' with ``metastable'' in condition
(i). Following FlowMM, FlowLLM, and ADiT, these are the \emph{hard}
metrics for crystal generation used in our comparisons.

\subsection{Protein Generation Metrics}
\label{app:metrics_protein}

The two protein evaluations use different sequence sources.
Table~\ref{tab:gen_prot} measures backbone designability with an
inverse-folding model, while Table~\ref{tab:performance} measures the
consistency of jointly generated sequences and backbones. We follow
the corresponding backbone-design and co-design task definitions of
ProteinBench~\citep{ye_proteinbench_2024}.

\paragraph{Backbone designability (Table~\ref{tab:gen_prot}).}
For a generated backbone $X\in\mathbb{R}^{L\times 3}$, ProteinMPNN
designs $8$ sequences $a_1,\ldots,a_8$. ESMFold~\citep{lin2024evolutionary} predicts a structure
for each sequence; let $\widehat{X}_k$ denote its C$\alpha$ coordinates.
The backbone's self-consistency error is
\[
\mathrm{scRMSD}_{\mathrm{bb}}(X)
=\min_{k=1,\ldots,8}\mathrm{RMSD}_{\mathrm{Kabsch}}
\bigl(X,\widehat{X}_k\bigr).
\]
Designability (``Des'', $\uparrow$) is the fraction of generated
backbones with $\mathrm{scRMSD}_{\mathrm{bb}}\leq2.0$\,\AA{}, following
\citet{watson_rfdiffusion_2023,yim_frameflow_2023}.

\paragraph{Diversity (``Div'', $\uparrow$).}
Number of structural clusters among the designable backbones, obtained
by single-linkage hierarchical clustering of pairwise TM-scores with
a $0.5$ cutoff. Higher values indicate more distinct generated folds.

\paragraph{DPT --- Designable Pairwise TM-score ($\downarrow$).}
Mean pairwise TM-score among the generated backbones that pass the
$2.0$\,\AA{} designability threshold. Lower values indicate greater
structural diversity within this subset. DPT compares generated
backbones with one another; scTM below compares a generated backbone
with the re-folded structure of its paired sequence.

\paragraph{Novelty (``Nov'', $\downarrow$).}
For each designable backbone $X$, we compute
$\max_{p\in\mathcal{R}}\mathrm{TM}(X,p)$, where $\mathcal{R}$ contains
reference structures from AFDB and PDB, following
\citet{li_proteinae_2025}. We report the mean over designable samples;
lower values indicate greater novelty relative to known folds.

\paragraph{Sequence--structure consistency (Table~\ref{tab:performance}).}
For each fixed length $L\in\{100,200\}$, \ours jointly generates
an amino-acid sequence $a$ and a C$\alpha$ backbone $X$. We pass
this generated sequence directly to ESMFold and extract the C$\alpha$
coordinates $\widehat{X}$ of its predicted structure:
\begin{gather*}
\widehat{X}=\operatorname{C}_{\alpha}\!\left[
\operatorname{ESMFold}(a)\right],\\
\mathrm{scTM}(X,a)=\mathrm{TM}(X,\widehat{X}),\qquad
\mathrm{scRMSD}(X,a)=\mathrm{RMSD}_{\mathrm{Kabsch}}(X,\widehat{X}).
\end{gather*}
TM-score measures structural agreement on a length-normalized scale,
with larger values indicating closer agreement. The RMSD uses optimal
Kabsch rotation and translation of corresponding C$\alpha$ atoms and
is reported in \AA{}. Both structures have the generated chain length
$L$. Each score evaluates the model's own sequence--backbone pair,
without ProteinMPNN redesign or best-of-8 sequence selection. Results
are reported separately for the two lengths.

\subsection{Latent-Space Analysis Metrics}
\label{app:metrics_latent}

\paragraph{Per-atom classical force-field energy (eV/atom, $\downarrow$).}
For molecule trajectories in Figure~\ref{fig:trajectory} we report the
per-structure energy computed with a classical force field (UFF/MMFF as
implemented in RDKit) evaluated on the decoder's intermediate
structures $(\mathbf{x}_t, \mathbf{a}_t)$. For crystal trajectories we
report the per-atom energy from the same CHGNet ML potential used for
stability evaluation above. These energies track the physical refinement
along the decoder's ODE path.

\section{Full Dataset Specification}
\label{app:datasets_full}

This appendix provides the per-domain dataset details abbreviated in
Section~\ref{subsec:datasets}.

\paragraph{Molecules.}
\textbf{QM9}~\citep{ramakrishnan2014quantum} consists of 130,000 stable
small organic molecules containing up to nine heavy atoms (C, N, O, F)
along with hydrogens. We follow the splits of
\citet{hoogeboom_equivariant_2022}.
\textbf{GEOM-Drugs}~\citep{axelrod_geom_2022} is a dataset of 1M
high-quality conformers of drug-like molecules; we use the splits of
\citet{vignac_midi_2023} and, following \citet{irwin_semlaflow_2025},
discard molecules with more than 72 heavy atoms from the training set
(accounting for 1\% of the data). At test time we sample molecule sizes
from the (unchanged) test-set size distribution.

\paragraph{Materials.}
\textbf{MP-20}~\citep{xie_crystal_2022} contains 45,231 metastable
crystal structures from the Materials
Project~\citep{jain_commentary_2013}, each with up to 20 atoms per unit
cell and spanning 89 elements; we use the CDVAE
splits~\citep{xie_crystal_2022}.
\textbf{QMOF}~\citep{rosen_machine_2021} is a database of 14,000 MOF
structures with quantum-chemical properties; we use the splits of
\citet{joshi_all-atom_nodate}.

\paragraph{Proteins.}
\textbf{AFDB-FS}~\citep{li_proteinae_2025} is a
filtered single-chain subset of the AlphaFold Protein Structure
Database, obtained via sequence-based MMseqs2 and structure-based
Foldseek~\citep{vankempen2024foldseek} clustering. It contains 588,318
structures with lengths in $[32, 256]$. For training we apply random
global SO(3) rotations as data augmentation. For reconstruction
evaluation we use the CASP14 and CASP15 benchmark sets.

\clearpage
\section{Method Design Rationale}
\label{app:method_rationale}

This appendix collects the design justifications abbreviated in the
main-paper architecture section.

\paragraph{Why a per-token latent instead of a pooled global code.}
Atomistic structures exhibit local geometric features (bond
environments, coordination shells) whose information content scales with
the number of atoms $N$; a fixed-size global latent can become restrictive
as $N$ grows. Our latent is
$\mathbf{z}\in\mathbb{R}^{N'\times d}$ with compact per-token dimension
$d$ (see Appendix~\ref{app:implementation_full}), which scales linearly with the number of sites. The encoder maps
$64$ hidden channels to $8$ latent channels for molecules/crystals
and $16$ for proteins, preserving one latent token per input site.

\paragraph{Why a non-equivariant Transformer.}
Equivariant layers encode physical symmetries as hard constraints, but
at substantial architectural-complexity and scaling cost. We instead
rely on random SO(3) rotation augmentation during training
(Section~\ref{subsec:featurization}) to induce rotation robustness, and
demonstrate empirically that this, combined with the iterative
flow-matching decoder, is sufficient to achieve near-lossless
reconstruction on molecules and crystals and sub-$0.03$\,\AA{} RMSD on
proteins.

\paragraph{Why output heads instead of stream-specific mid-backbone
cross-attention.}
Following Tabasco~\citep{vonessen_tabasco_2025}, we localise the
coordinate / atom-type specialisation in two parallel cross-attention
output heads rather than interleaving stream-specific cross-attention
inside the backbone. This keeps every Transformer block in the backbone
fully shared across its constituent structural categories, and places
all cross-attention compute at the output, where it directly conditions
the two prediction heads on the embedded noisy inputs.

\paragraph{Why we do not use self-conditioning.}
The implementation uses cross-attention output heads without
self-conditioning~\citep{chen2023analog}. This keeps the decoder
input interface unchanged across iterative steps.

\section{KL Regularisation and Training Weights}
\label{app:training_objective_full}

The coordinate and type objectives are given in
Eqs.~\ref{eq:loss_coord} and~\ref{eq:loss_atom}. This appendix supplies
the KL term that completes Eq.~\ref{eq:loss_total}, together with the
training weights.

\paragraph{KL regularisation.}
Unlike DiTo~\citep{chen_diffusion_2025}, we pair a stochastic encoder
with the decoder and add a light KL term pushing
$q_\phi(\mathbf{z}\mid\mathcal{G})$ towards
$\mathcal{N}(\mathbf{0},\mathbf{I})$:
\begin{equation}
    \mathcal{L}_{\text{KL}} =
    D_{\mathrm{KL}}\bigl(q_\phi(\mathbf{z}\mid\mathcal{G}) \,\|\,
    \mathcal{N}(\mathbf{0},\mathbf{I})\bigr) =
    \tfrac{1}{2}\sum_{j}\bigl(\mu_j^2 + \sigma_j^2 -
    \log\sigma_j^2 - 1\bigr).
    \label{eq:loss_kl}
\end{equation}
The latent flow model learns the encoded-data distribution separately
(Section~\ref{subsec:latent_fm}).

Weight values ($\lambda_{\text{atom}}{=}0.1$,
$\lambda_{\text{KL}}{=}10^{-6}$) are listed alongside the rest of the
training configuration in Appendix~\ref{app:implementation_full}.

\section{Full Implementation Details}
\label{app:implementation_full}

This appendix gives the complete architectural, training, and evaluation
specification for the \ours models summarised in
Section~\ref{subsec:datasets}.

\subsection{Architecture}
\label{app:impl_arch}

The autoencoder uses a pre-norm vanilla Transformer with hidden
dimension $d_{\text{model}}{=}64$ and $8$ attention heads throughout. No
equivariant layer, message-passing operator, graph attention, or
pretrained discriminator is used anywhere in the architecture.
\begin{itemize}
  \item \textbf{Joint molecule--crystal autoencoder.} Encoder depth
  $L_{\text{enc}}{=}4$, decoder depth $L_{\text{dec}}{=}16$, per-token
  latent dimension $d{=}8$ (so
  $\mathbf{z}\in\mathbb{R}^{N'\times 8}$).
  \item \textbf{Joint molecule--crystal latent DiT.} $L_{\text{dit}}{=}24$
  layers, $d_{\text{model}}{=}1024$, $16$ attention heads.
  \item \textbf{Protein autoencoder.} Encoder depth $L_{\text{enc}}{=}4$,
  decoder depth $L_{\text{dec}}{=}16$, per-token latent dimension
  $d{=}16$.
  \item \textbf{Protein latent DiT prior.} $L_{\text{dit}}{=}15$ layers,
  $d_{\text{model}}{=}768$, $12$ attention heads.
\end{itemize}

\subsection{Encoder and Decoder Details}
\label{app:encoder_decoder_details}

\paragraph{Encoder.}
The encoder $\phi$ maps a sequence of $N'$ input tokens to a sequence
of latent tokens $\mathbf{z}\in\mathbb{R}^{N'\times d}$, where $N'=N$
for aperiodic systems and $N'=N{+}3$ for periodic systems (the three
trailing tokens encode the lattice vectors). A per-token (rather than
pooled) latent preserves per-atom granularity: local geometric
information scales with $N$, so a fixed-size global bottleneck can
become restrictive as system size increases. We use a standard pre-norm
Transformer with $L_{\text{enc}}$ layers and multi-head self-attention;
after the final layer, two parallel linear heads produce a Gaussian
posterior
$\mathbf{z}_i = \boldsymbol{\mu}_i + \boldsymbol{\sigma}_i \odot
\boldsymbol{\varepsilon}_i$,
$\boldsymbol{\varepsilon}_i\sim\mathcal{N}(\mathbf{0},\mathbf{I})$. We
use a small latent channel $d$ (see
Appendix~\ref{app:implementation_full}), keeping the total latent size
compact while preserving enough information for near-exact decoding.

\paragraph{Flow-matching decoder.}
The decoder $\psi$ is a non-equivariant Transformer that reconstructs
$(\mathbf{X}, \mathbf{A})$ from $\mathbf{z}$ conditioned on partially
noised inputs at time $t\in[0,1]$. We follow DiTo's~\citep{chen_diffusion_2025}
single-objective design but replace its score-matching formulation with
a continuous conditional flow-matching objective for coordinates and a
discrete flow-matching objective for atom types. The decoder receives
noisy $(\mathbf{X}_t, \mathbf{A}_t)$ projected through separate
linears, summed with a sinusoidal time embedding and a learnable
positional embedding:
\begin{equation}
    \mathbf{h}_i^{(0)} = W_x\,\mathbf{x}_{t,i} + W_A\,\mathbf{a}_{t,i}
    + \mathbf{e}^{\text{time}}(t) + \mathbf{e}^{\text{pos}}_i,
    \label{eq:dec_input}
\end{equation}
where $W_x\in\mathbb{R}^{d_{\mathrm{model}}\times 3}$ and
$W_A\in\mathbb{R}^{d_{\mathrm{model}}\times K}$; categorical labels
are represented as one-hot vectors $\mathbf{a}_{t,i}$.
The joint sequence $[\mathbf{h}^{(0)};\, \mathbf{z}]$ (concatenated
along the sequence dimension, yielding $2N'$ tokens) is processed by
$L_{\text{dec}}$ pre-norm Transformer
blocks~\citep{vaswani2017attention}. Following the two-stream output
design of Tabasco~\citep{vonessen_tabasco_2025}, two parallel
cross-attention output heads, one for coordinates and one for atom
types, use the final hidden states as queries and the embedded
noisy-input tokens $\{\mathbf{c}_i\}$ as keys/values to produce
$\hat{\mathbf{X}}_1^\theta$ and $\hat{\mathbf{A}}_1^\theta$
(Figure~\ref{fig:main}b). No domain-specific attention, equivariant
layer, or self-conditioning~\citep{chen2023analog} is used in the
backbone; random SO(3) augmentation at training time provides rotation
robustness (Appendix~\ref{app:method_rationale}).

\subsection{Training}
\label{app:impl_training}

All stages use Adam (autoencoder) and AdamW (latent prior) with a
constant learning rate of $1{\times}10^{-4}$, weight decay $0$, no
warm-up, and no scheduler. The joint molecule--crystal autoencoder is
trained with batch size $2048$ (each batch drawn $50$/$50$ from QM9 and
MP-20); the protein autoencoder uses batch size $256$.

The total reconstruction loss follows Eq.~\ref{eq:loss_total} with the
fixed weights
\[
\lambda_{\text{coord}}{=}1.0,\quad
\lambda_{\text{atom}}{=}0.1,\quad
\lambda_{\text{KL}}{=}1{\times}10^{-6}.
\]
The flow-matching schedule uses a lower time cutoff
$t_{\min}{=}10^{-2}$; no additional $\sigma_{\min}$ is imposed.

The joint molecule--crystal model was trained on a single GPU for
approximately $48$ GPU-hours.

\subsection{Evaluation Protocol}
\label{app:impl_eval}

\paragraph{Reconstruction RMSD.}
All reconstruction RMSDs are computed after Kabsch SVD alignment between
the decoded and reference coordinates. Protein CA-RMSD uses the same
Kabsch pipeline on the $C_\alpha$ trace.

\paragraph{Crystal Structure-Match Rate.}
Both crystal generation and crystal reconstruction structure-match rates
use the same \texttt{pymatgen.StructureMatcher} pipeline with the
standard thresholds $\mathrm{ltol}{=}0.3$, $\mathrm{stol}{=}0.5$,
$\mathrm{angle\_tol}{=}10^{\circ}$~\citep{xie_crystal_2022}; we
additionally report Kabsch-aligned RMSD for reconstruction as a
continuous fidelity metric.

\paragraph{Latent-sampling and decoder NFE.}
For joint molecule--crystal generation, the DiT latent prior uses
$\mathrm{NFE}_{\mathrm{prior}}=100$ Euler steps, followed by
$\mathrm{NFE}_{\mathrm{dec}}=10$ Euler steps in the flow-matching
decoder. Protein latent sampling uses
$\mathrm{dt}_{\mathrm{latent}}=2.5\times10^{-3}$
($\mathrm{NFE}_{\mathrm{prior}}=400$); its flow-matching decoder
also uses $\mathrm{NFE}_{\mathrm{dec}}=10$.
For the joint model, classifier-free guidance combines conditional
and unconditional predictions at each latent step. The reported prior
NFE counts evaluations of this guided vector field.

\section{Round-Trip Physical Diagnostics}
\label{app:roundtrip}

Table~\ref{tab:roundtrip} compares reference structures with their
autoencoder reconstructions. Molecular strain is the UFF energy difference in RDKit between
the supplied conformation and its force-field-relaxed counterpart.
Atom stability measures valid valency under the bond-inference
convention; molecule stability requires all atoms to pass. These are
fractions in Table~\ref{tab:roundtrip}, distinct from RDKit validity
and PoseBusters validity.

The crystal diagnostic uses $100$ MP-20 training structures evaluated
under the same energy procedure before and after reconstruction.
Table~\ref{tab:roundtrip} records validity, mean/std.\ energy above
hull, stable counts, and the displacement under relaxation, measuring
the physical changes introduced by reconstruction.

\section{Inference Profiling and Decoder NFE}
\label{app:profiling}

Tables~\ref{tab:throughput} and~\ref{tab:decoder_time} collect the
inference profiles; all models are timed on the same single GPU.
Throughput is amortised per sample; component timings
are measured per call. The component profile uses the same ten-step
flow-matching decoder as the generation protocol
(Appendix~\ref{app:impl_eval}).

\begin{table}[H]
\centering
\small
\setlength{\tabcolsep}{7pt}
\caption{\textbf{End-to-end inference profiles on a single GPU.}
Seconds per sample; 10k times are the originally reported values,
rounded independently. Protein values are generation times.}
\label{tab:throughput}
\begin{tabular}{llrrl}
\toprule
Domain & Model & Params (M) & s/sample & 10k samples \\
\midrule
QM9 & SemlaFlow & 23.30 & 0.0353 & 0.098 h \\
& ADiT & 508.12 & 0.0536 & 0.149 h \\
& \ours & 458.26 & 0.0553 & 0.154 h \\
& EQGAT-diff & 3.61 & 0.1220 & 0.339 h \\
\midrule
MP-20 & \ours & 458.26 & 0.0870 & 0.242 h \\
& FlowMM & 24.48 & 0.6898 & 1.916 h \\
& MatterGen & 44.56 & 2.5755 & 7.154 h \\
\midrule
Protein & \ours & 318.0 & 0.634 & 1 h 45 min \\
& DPLM2-650M & 673.4 & 10.319 & 28 h 39 min \\
& RFdiffusion & 59.8 & 27.296 & 75 h 49 min \\
\bottomrule
\end{tabular}
\end{table}

\begin{table}[H]
\centering
\small
\setlength{\tabcolsep}{12pt}
\caption{\textbf{QM9 component profile.} Call-level times; decoder
share is decode time divided by latent sampling plus decoding time.}
\label{tab:decoder_time}
\begin{tabular}{lrr}
\toprule
Metric & \ours & ADiT \\
\midrule
Latent sampling (ms) & 3456.22 & 3431.76 \\
Decoding (ms) & 25.65 & 3.62 \\
Decoder share (\%) & 0.74 & 0.11 \\
Decoder parameters (M) & 1.22 & 51.09 \\
Decoder forward passes & 10 & 1 \\
\bottomrule
\end{tabular}
\end{table}

\begin{figure}[H]
    \centering
    \begin{minipage}[c]{0.60\linewidth}
    \includegraphics[width=\linewidth]{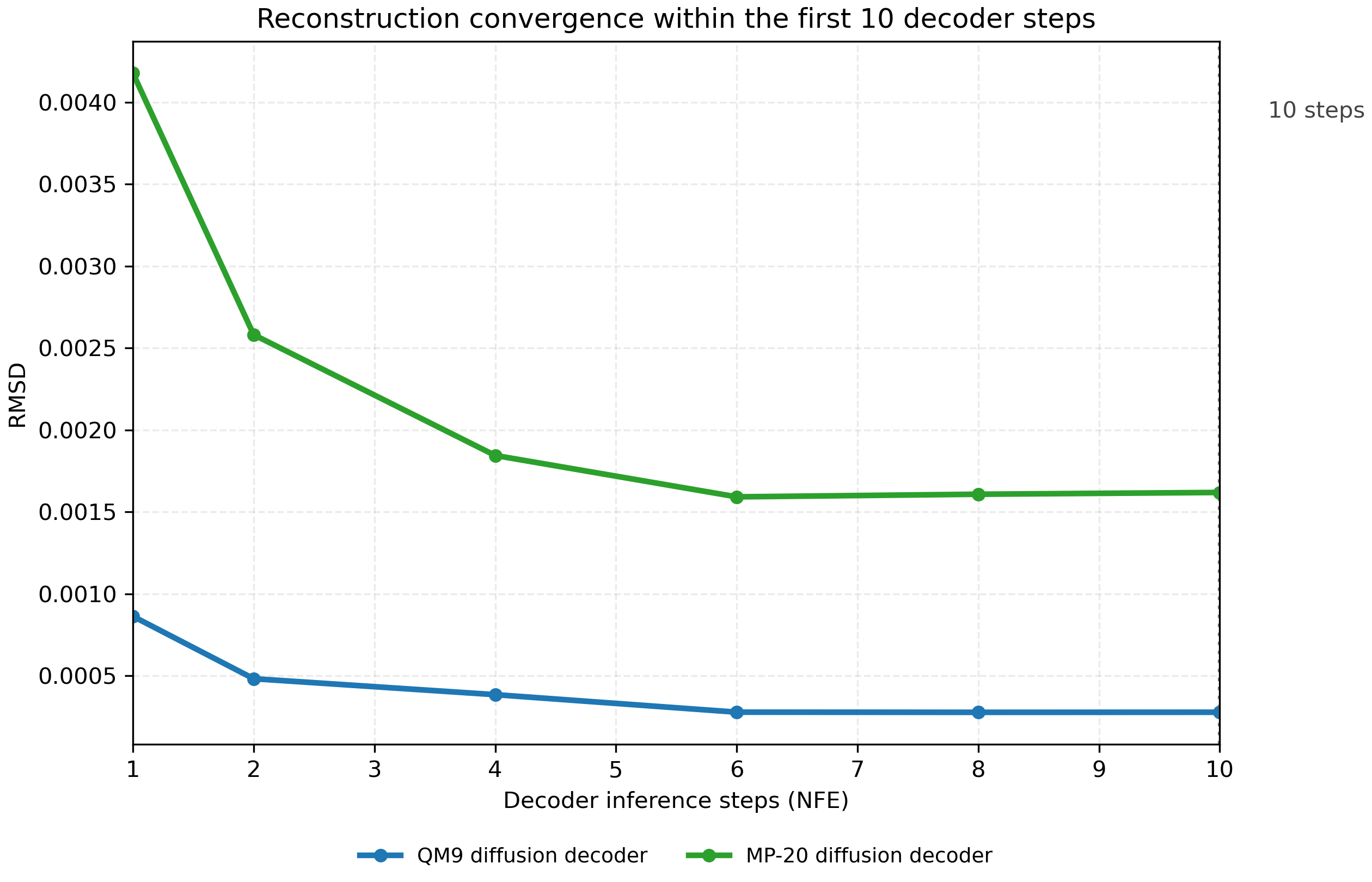}
    \end{minipage}\hfill
    \begin{minipage}[c]{0.36\linewidth}
    \caption{\textbf{Reconstruction fidelity vs decoder NFE.} RMSD on
    QM9 (blue) and MP-20 (green) as a function of the number of decoder
    function evaluations. Both curves drop sharply between NFE${=}1$ and
    NFE${=}4$ and saturate by NFE${\approx}6$.}
    \label{fig:nfe}
    \end{minipage}
\end{figure}

\clearpage
\paragraph{Reconstruction versus decoder NFE.}
Figure~\ref{fig:nfe} shows reconstruction fidelity across decoder step counts.

\section{t-SNE of the Joint Latent Space}
\label{app:tsne}

\begin{figure}[H]
    \centering
    \includegraphics[width=0.7\linewidth]{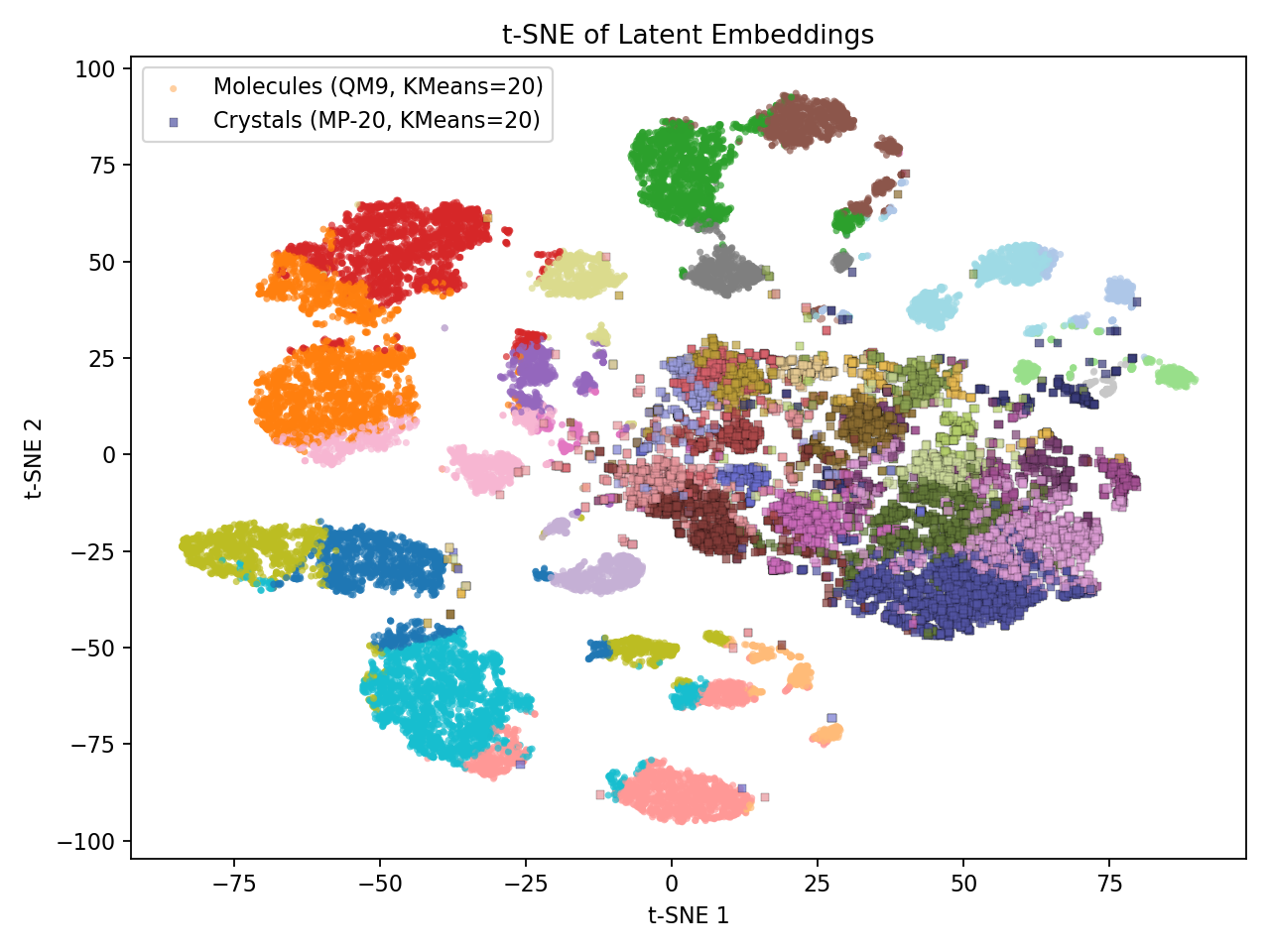}
    \caption{\textbf{t-SNE of \ours latents.} QM9 (circles) and MP-20
    (squares) separate without explicit domain conditioning; colours are
    within-domain KMeans clusters.}
    \label{fig:tsne}
\end{figure}

Figure~\ref{fig:tsne} projects the joint molecule--crystal encoder
outputs into two dimensions. QM9 molecules and MP-20 crystals occupy
largely separated regions, and colours denote within-domain KMeans
clusters.

\clearpage
\section{Within-Domain Latent-Space Interpolation}
\label{app:interpolation}

\begin{figure}[H]
    \centering
    \includegraphics[width=0.80\linewidth]{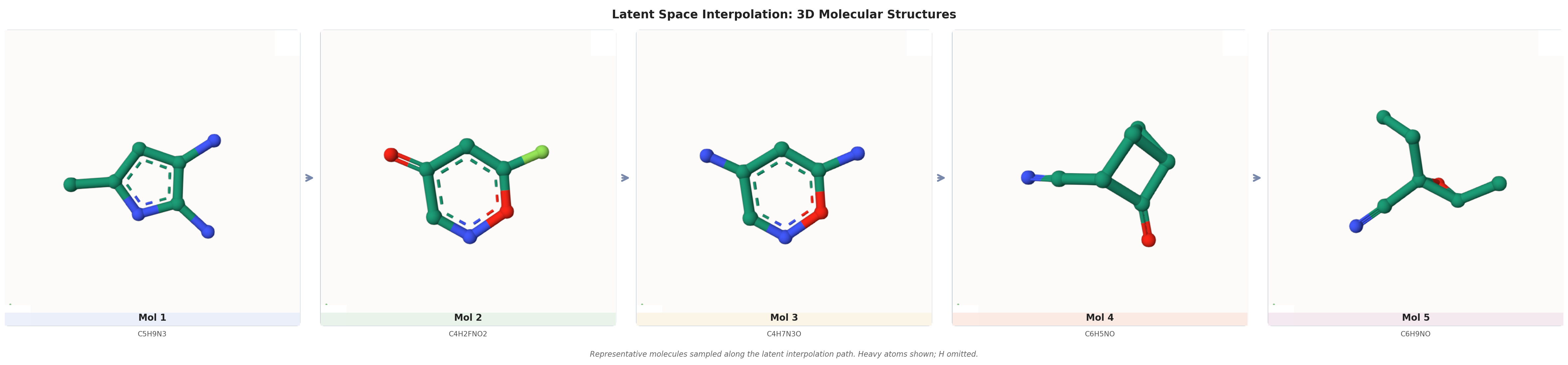}\\[2pt]
    \includegraphics[width=0.80\linewidth]{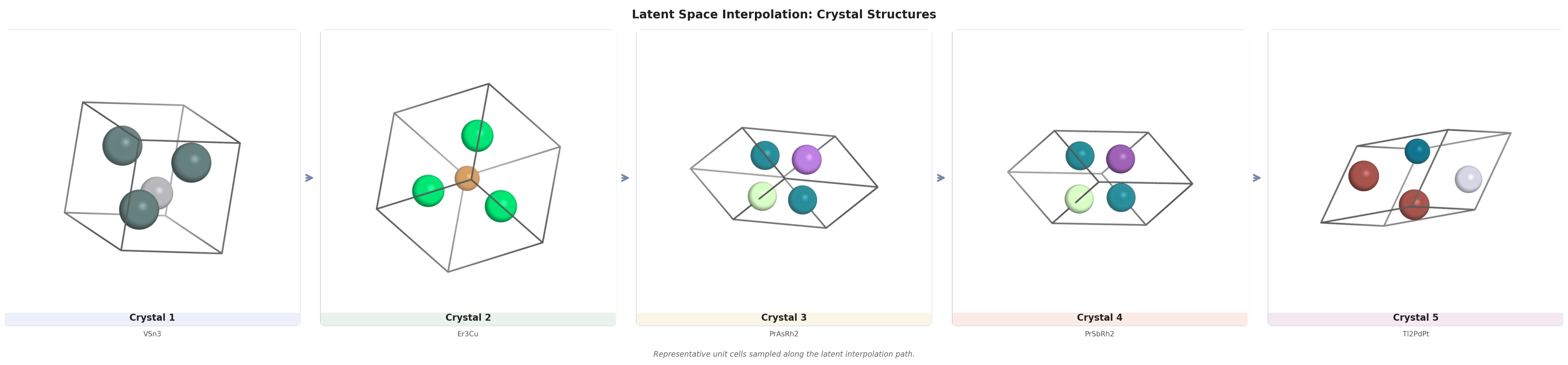}\\[2pt]
    \includegraphics[width=0.80\linewidth]{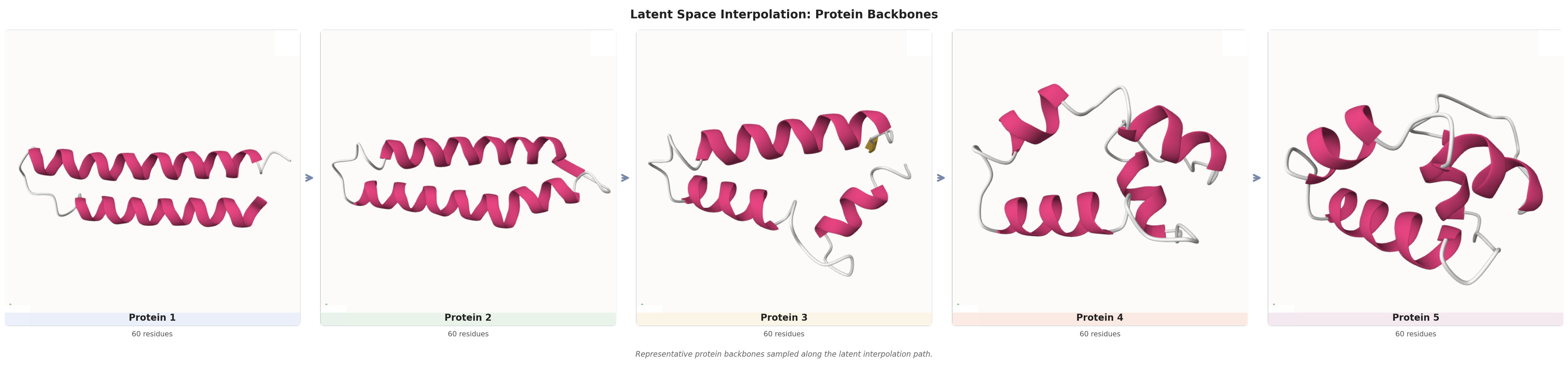}
    \caption{\textbf{Within-domain latent-space interpolation.}
    Structures decoded along straight-line interpolations between pairs
    of \ours latents: molecules (top), crystals (middle), proteins
    (bottom).}
    \label{fig:interp}
\end{figure}

Figure~\ref{fig:interp} shows structures decoded along straight-line
paths between pairs of encoded structures, separately within each
domain. The examples illustrate gradual changes in molecular shape,
crystal cells, and protein backbones.

\section{Protein Reconstruction: Means and Standard Deviations}
\label{app:recon_prot_full}

Table~\ref{tab:recon_prot_full} reproduces the protein-backbone
C$\alpha$ reconstruction RMSDs from Table~\ref{tab:recon_prot}
with their mean$\,\pm\,$std.\ values. Baseline values come from
Table~1 of \citet{li_proteinae_2025}; \ours values are from our
evaluation. Standard deviations are calculated across structures.

\begin{table}[H]
    \centering
    \small
    \setlength{\tabcolsep}{4pt}
    \caption{\textbf{Protein C$\alpha$ reconstruction RMSD --- full table.}
Mean$\,\pm\,$std.\ in \AA{} ($\downarrow$), with the same evaluation
settings as Table~\ref{tab:recon_prot}. \ours values are bold.}
    \label{tab:recon_prot_full}
    \begin{tabular}{lccccc}
        \toprule
        & \multicolumn{3}{c}{CASP14} & \multicolumn{2}{c}{CASP15} \\
        \cmidrule(lr){2-4} \cmidrule(lr){5-6}
        Method & T & T-dom & oligo & TS-dom & oligo \\
        \midrule
        CHEAP & 11.15\,$\pm$\,9.88 & 8.99\,$\pm$\,9.18 & 9.93\,$\pm$\,10.55 & 10.22\,$\pm$\,11.23 & 9.22\,$\pm$\,12.57 \\
        ESM3 VQ-VAE & 1.02\,$\pm$\,1.82 & 0.66\,$\pm$\,0.42 & 3.08\,$\pm$\,7.39 & 1.23\,$\pm$\,1.26 & 1.94\,$\pm$\,2.43 \\
        DPLM-2$^{*}$ & 1.99\,$\pm$\,2.03 & 1.87\,$\pm$\,1.78 & 2.70\,$\pm$\,5.05 & 3.31\,$\pm$\,5.69 & 3.50\,$\pm$\,6.23 \\
        ProteinAE & 0.23\,$\pm$\,0.15 & 0.22\,$\pm$\,0.11 & 0.31\,$\pm$\,0.22 & 0.28\,$\pm$\,0.20 & 0.37\,$\pm$\,0.50 \\
        \midrule
        \textbf{\ours} & \textbf{0.010\,$\pm$\,0.005} & \textbf{0.011\,$\pm$\,0.004} & \textbf{0.013\,$\pm$\,0.008} & \textbf{0.013\,$\pm$\,0.009} & \textbf{0.014\,$\pm$\,0.005} \\
        \bottomrule
    \end{tabular}
\end{table}

\clearpage
\section{Protein Performance Across the Evaluated Length Range}
\label{app:ablation_length}

\begin{figure}[H]
    \centering
    \includegraphics[width=0.85\linewidth]{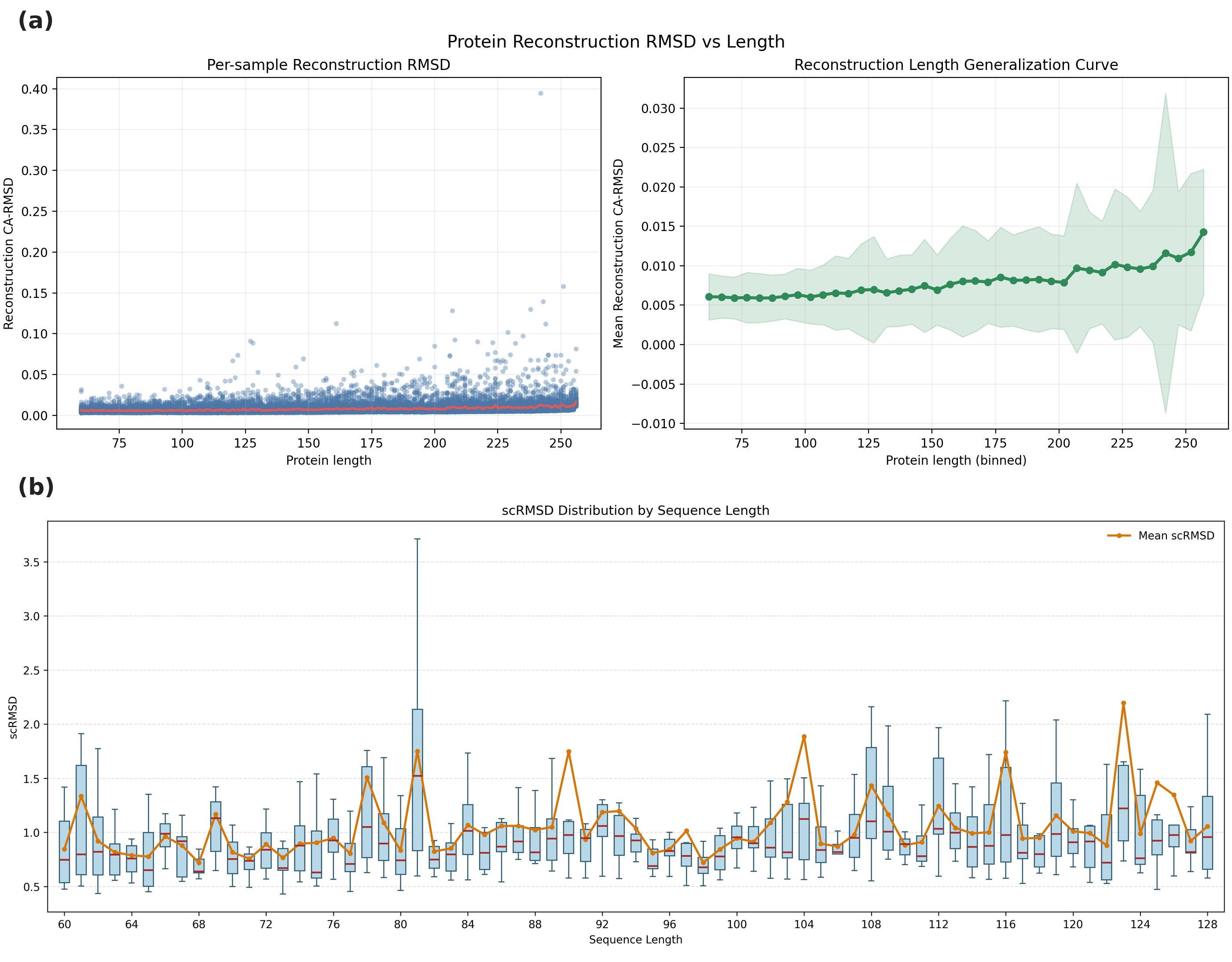}
    \caption{\textbf{Protein metrics across evaluated lengths.} \textbf{(a)}
    Per-sample reconstruction CA-RMSD vs protein length on AFDB-FS; the
    binned mean (right) stays below $0.015$\,\AA{} across the full
    60--260 residue range. \textbf{(b)} Distribution of generation
    scRMSD as a function of generated sequence length; the mean (orange)
    is flat across 60--128 residues.}
    \label{fig:length}
\end{figure}

Figure~\ref{fig:length} describes reconstruction and generation as a
function of protein length. The reconstruction plot spans approximately
$60$--$260$ residues, with the binned mean rising from about
$0.006$ to $0.014$\,\AA{}. The generation plot spans $60$--$128$
residues with approximately constant mean scRMSD.

\clearpage
\section{Material Generation: Full MP-20 Breakdown}
\label{app:gen_mat_full}

Table~\ref{tab:gen_mat_full} provides the full MP-20 validity,
metastability, and stability breakdown for the main comparison.
Metric definitions are in Appendix~\ref{app:metrics_crystal}.

\begin{table}[H]
    \centering
    \small
    \setlength{\tabcolsep}{5pt}
    \caption{\textbf{Material generation on MP-20 --- full breakdown.}
    All columns in \%, higher is better. ``--'' denotes numbers not
    reported by the original authors.}
    \label{tab:gen_mat_full}
    \begin{tabular}{lccccccc}
        \toprule
        & \multicolumn{3}{c}{Validity Rate} & Metastable & Stable & M.S.U.N. & S.U.N. \\
        Model & Structure & Composition & Overall & rate & rate & rate & rate \\
        \midrule
        CDVAE          & 100.00 & 86.70 & --    & --    & 1.6  & --    & --   \\
        DiffCSP        & 100.00 & 83.25 & --    & --    & 5.0  & --    & 3.3  \\
        UniMat         & 97.2   & 89.4  & --    & --    & --   & --    & --   \\
        FlowMM         & 96.85  & 83.19 & 80.30 & 30.6  & 4.6  & 22.5  & 2.8  \\
        FlowLLM        & 99.94  & 90.84 & 90.81 & 66.9  & 13.9 & 26.3  & 4.7  \\
        Jointly-trained ADiT & 99.74 & 92.14 & 91.92 & 81.0 & 15.4 & 28.2 & 5.3 \\
        \midrule
        \textbf{\ours (MP20-only)} & 98.29 & 88.89 & 88.40 & 79.6 & 14.1 & 24.3 & 4.6 \\
        \textbf{\ours (joint)}     & 99.29 & 90.49 & 90.02 & 83.0 & 15.8 & 30.2 & 5.4 \\
        \bottomrule
    \end{tabular}
\end{table}

Table~\ref{tab:gen_mat_extended} adds five crystal-specific baselines. SGFM achieves S.U.N.\ $6.9\%$ and M.S.U.N.\
$30.3\%$; \ours achieves $5.4\%$ and $30.2\%$, respectively,
while supporting molecules and protein backbones with the same
autoencoding architecture.

\begin{table}[H]
    \centering
    \small
    \setlength{\tabcolsep}{5pt}
    \caption{\textbf{Extended crystal-generation comparison on MP-20 (\%).}
Higher is better; ``--'' denotes unavailable values.
MatterGen and Crystalite values follow Appendix~F.1, Table~5 of
\citet{veljkovic2026crystalite}, using its MLIP-based MatterGen evaluation
pipeline; its $0.1$\,eV/atom SUN rate is shown as M.S.U.N.
OMatG, SGFM, and SGEquiDiff values follow Table~5 of \citet{puny2025sgfm}.}
    \label{tab:gen_mat_extended}
    \begin{tabular}{lccccc}
        \toprule
        Model & Struct\,$\uparrow$ & Comp\,$\uparrow$ & Stable\,$\uparrow$ & M.S.U.N.\,$\uparrow$ & S.U.N.\,$\uparrow$ \\
        \midrule
        CDVAE$^\dagger$  & 100.00 & 86.70 & 1.6  & --    & --   \\
        DiffCSP$^\dagger$  & 100.00 & 83.25 & 5.0  & --    & 3.3  \\
        UniMat$^\dagger$   & 97.2   & 89.4  & --   & --    & --   \\
        FlowMM         & 96.85  & 83.19 & 4.6  & 22.5  & 2.8  \\
        FlowLLM        & 99.94  & 90.84 & 13.9 & 26.3  & 4.7  \\
        Jointly-trained ADiT & 99.74 & 92.14 & 15.4 & 28.2 & 5.3 \\
        MatterGen$^{\ddagger}$ & 100.00 & 83.48 & -- & 23.75 & -- \\
        OMatG$^{\ddagger}$ & 95.05 & 82.84 & -- & 23.7 & 2.2 \\
        Crystalite$^{\ddagger}$ & 100.00 & 84.62 & -- & 24.26 & -- \\
        SGFM (empirical)$^{\ddagger}$ & 99.87 & 86.81 & -- & 30.3 & 6.9 \\
        SGEquiDiff$^{\ddagger}$ & 99.25 & 86.16 & -- & 25.8 & -- \\

        \midrule
        \textbf{\ours (MP20-only)} & 98.29 & 88.89 & 14.1 & 24.3 & 4.6 \\
        \textbf{\ours (joint)}     & 99.29 & 90.49 & 15.8 & 30.2 & 5.4 \\
        \bottomrule
        \multicolumn{6}{l}{\scriptsize $^\dagger$\,CDVAE~\citep{xie_crystal_2022},
        DiffCSP~\citep{jiao_crystal_nodate},
        UniMat~\citep{yang_unimat_2024}.}\\
        \multicolumn{6}{l}{\scriptsize $^{\ddagger}$\,MatterGen~\citep{zeni2025mattergen}, OMatG~\citep{hoellmer2025omatg}, Crystalite~\citep{veljkovic2026crystalite},}\\
        \multicolumn{6}{l}{\scriptsize SGFM~\citep{puny2025sgfm}, SGEquiDiff~\citep{chang2025sgequidiff}.}
    \end{tabular}
\end{table}

\clearpage
\section{Molecule Generation: Per-PoseBusters-Check Breakdown}
\label{app:gen_mol_pb_full}

Table~\ref{tab:gen_mol_pb_full} reports the per-check pass rates for
each of the seven PoseBusters sanity
checks~\citep{buttenschoen2024posebusters} for \ours on QM9 (joint
model) and GEOM-Drugs ($10$\,k samples each), together with the
aggregate \emph{PB Valid} rate (passes all seven checks) reported in
Tables~\ref{tab:gen_mol_qm9}--\ref{tab:gen_mol_geom}. The breakdown
shows that on QM9 the residual is dominated by the \emph{internal
energy} check ($96.8\%$), and on GEOM-Drugs by \emph{internal steric
clash} ($96.7\%$) and \emph{bond angles} ($97.8\%$); other checks are
at or near $100\%$.

\begin{table}[H]
\centering
\small
\setlength{\tabcolsep}{6pt}
\caption{\textbf{Per-PoseBusters-check pass rates (\%, $\uparrow$) for
\ours on QM9 and GEOM-Drugs.} Last row is the aggregate
\emph{PoseBusters Valid} rate (passes all seven checks), corresponding
to the PB Valid column in
Tables~\ref{tab:gen_mol_qm9}--\ref{tab:gen_mol_geom}.}
\label{tab:gen_mol_pb_full}
\begin{tabular}{lcc}
\toprule
Test (\% pass) & \ours (QM9, joint) & \ours (GEOM-Drugs) \\
\midrule
All atoms connected         & 100.00 & 99.48 \\
Bond lengths                & 99.59  & 99.07 \\
Bond angles                 & 100.00 & 97.84 \\
No internal steric clash    & 99.90  & 96.70 \\
Aromatic ring flatness      & 100.00 & 100.00 \\
Double bond flatness        & 99.90  & 99.90 \\
Reasonable internal energy  & 96.83  & 98.45 \\
\midrule
\textbf{PoseBusters Valid (all 7)} & \textbf{96.52} & \textbf{92.58} \\
\bottomrule
\end{tabular}
\end{table}

\section{MOF Generation: Full MOFChecker Sanity Checks}
\label{app:mof_gen_full}

Table~\ref{tab:mof_gen} reports the full set of MOFChecker sanity
checks for 1,000 unconditionally sampled MOF structures from \ours
(trained on QMOF) and the jointly-trained ADiT baseline. The main text
reports MOF reconstruction in Table~\ref{tab:recon_nonprot}; this
appendix provides the generation-side comparison. \ours outperforms ADiT
by $+6.1$ points on the aggregate validity rate ($16.28\%$ vs
$10.2\%$), with consistent improvements on nearly every individual
sanity check.

\begin{table}[H]
\centering
\caption{\textbf{MOF generation benchmark (full MOFChecker breakdown).}
Sanity checks from MOFChecker for 1,000 sampled structures are reported
for \ours and Joint ADiT. Higher values are better for metrics marked
with $\uparrow$, lower values are better for those marked with
$\downarrow$.}
\label{tab:mof_gen}
\begin{tabular}{lcc}
\toprule
Test (\%) & \ours & Joint ADiT \\
\midrule
Has carbon $\uparrow$ & 100.0 & 100.0 \\
Has hydrogen $\uparrow$ & 99.8 & 100.0 \\
Has atomic overlap $\downarrow$ & 5.2 & 10.8 \\
Has overcoord.\ C $\downarrow$ & 12.8 & 34.3 \\
Has overcoord.\ N $\downarrow$ & 0.4 & 1.6 \\
Has overcoord.\ H $\downarrow$ & 1.3 & 3.6 \\
Has undercoord.\ C $\downarrow$ & 39.2 & 72.1 \\
Has undercoord.\ N $\downarrow$ & 26.9 & 39.9 \\
Has undercoord.\ rare earth $\downarrow$ & 0.2 & 0.8 \\
Has metal $\uparrow$ & 100.0 & 99.4 \\
Has lone molecule $\downarrow$ & 49.8 & 83.2 \\
Has high charge $\downarrow$ & 0.2 & 2.5 \\
Has suspicious terminal oxo $\downarrow$ & 3.1 & 5.8 \\
Has undercoord.\ alkali $\downarrow$ & 0.7 & 6.4 \\
Has geom.\ exposed metal $\downarrow$ & 4.0 & 9.6 \\
\midrule
Validity rate (all passed) $\uparrow$ & \textbf{16.28} & 10.2 \\
\bottomrule
\end{tabular}
\end{table}

\clearpage
\section{Generated Sample Visualisations}
\label{app:samples_full}

\begin{figure}[!ht]
    \centering
    \includegraphics[width=0.95\linewidth,height=0.72\textheight,keepaspectratio]{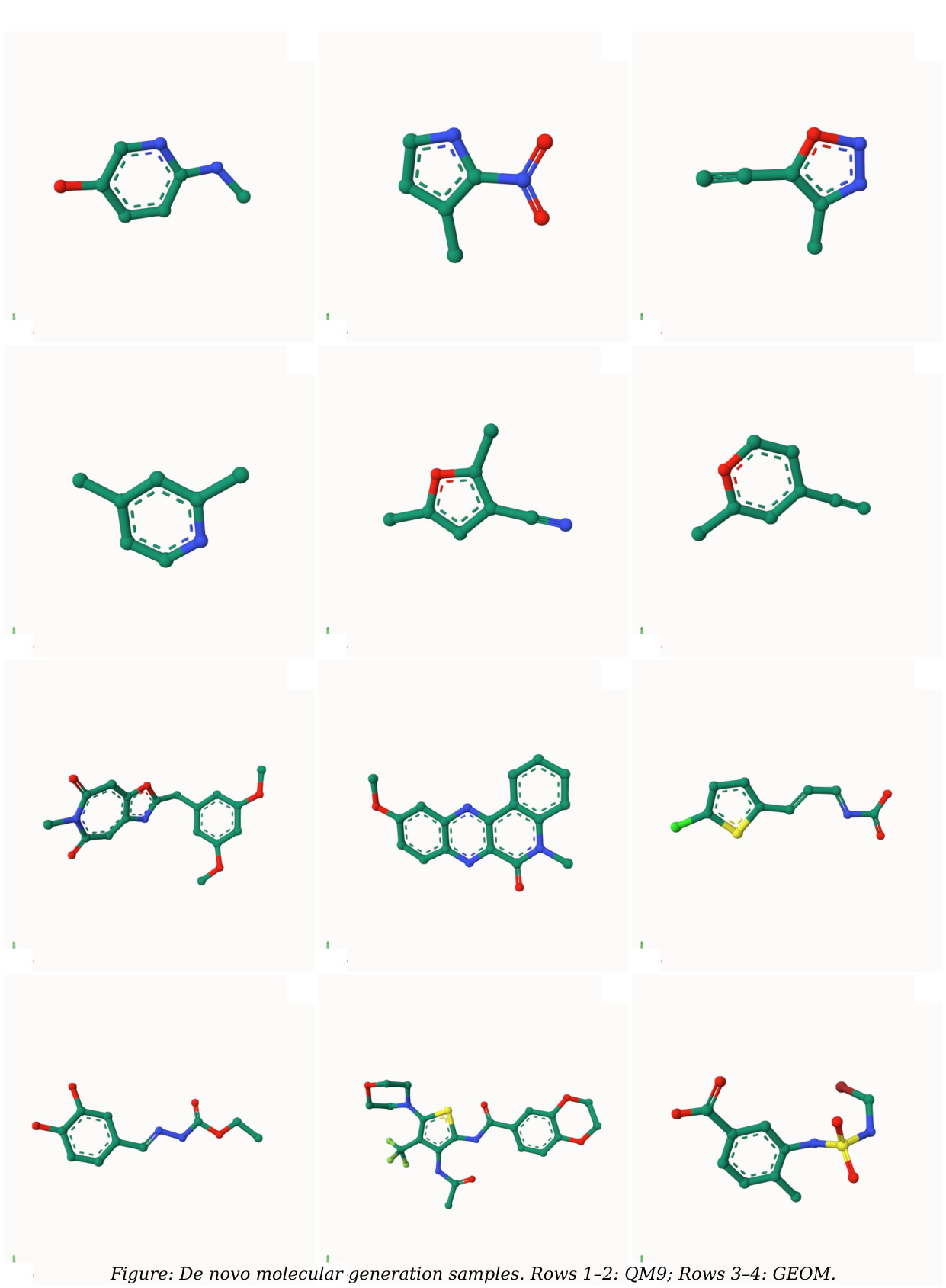}
    \caption{\textbf{Generated samples on QM9 and GEOM-Drugs.} Top:
    unconditional samples from the joint molecule + crystal \ours prior
    (QM9). Bottom: unconditional samples from the GEOM-Drugs \ours
    prior.}
    \label{fig:samples_qm9}
\end{figure}

\begin{figure}[!t]
    \centering
    \includegraphics[width=0.95\linewidth,height=0.82\textheight,keepaspectratio]{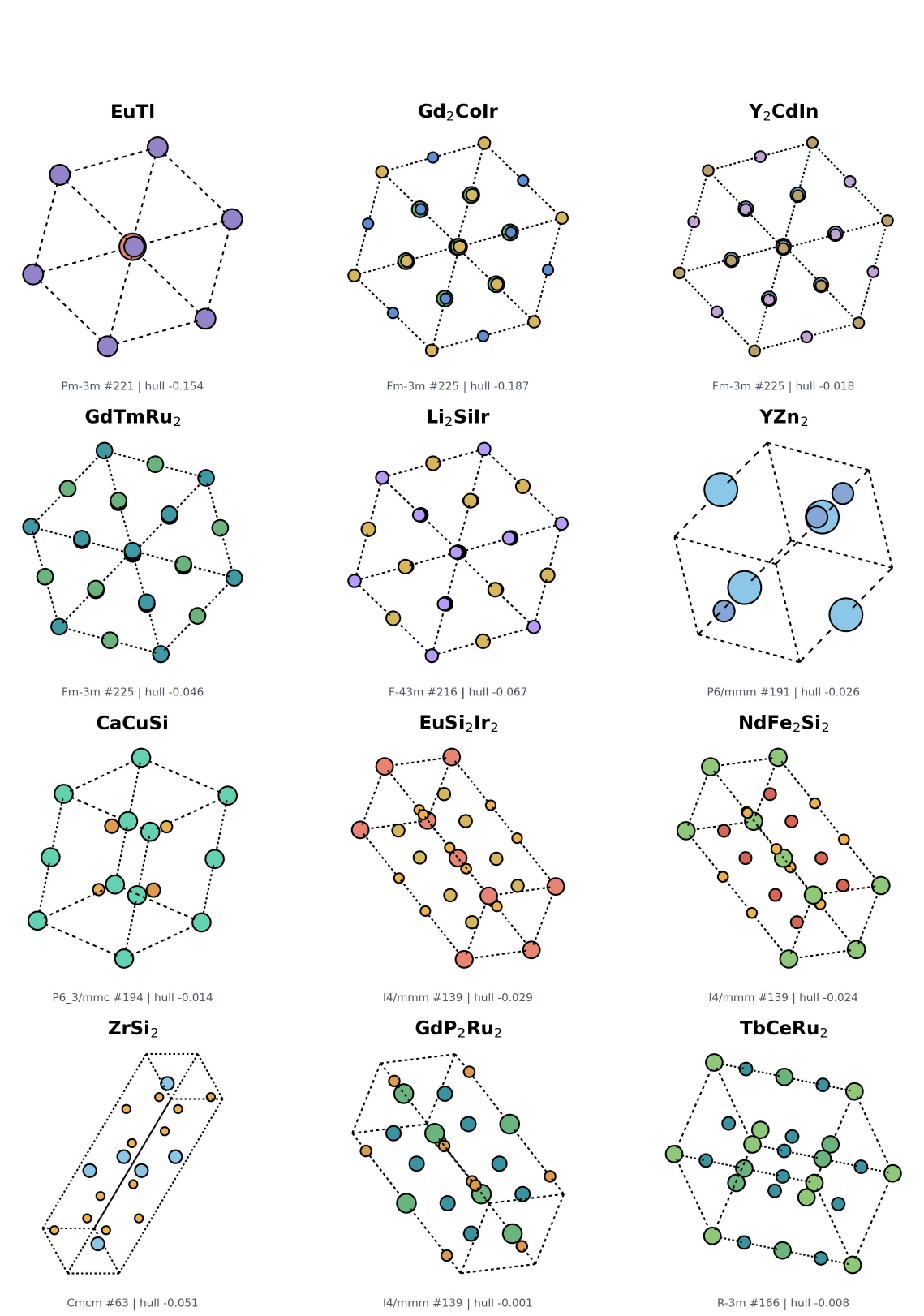}
    \caption{\textbf{Generated samples on MP-20.} Unconditional crystal
    samples from the joint molecule + crystal \ours prior, decoded with
    their lattice tokens to inorganic crystal structures.}
    \label{fig:samples_mp20}
\end{figure}

\begin{figure}[!t]
    \centering
    \includegraphics[width=0.95\linewidth,height=0.82\textheight,keepaspectratio]{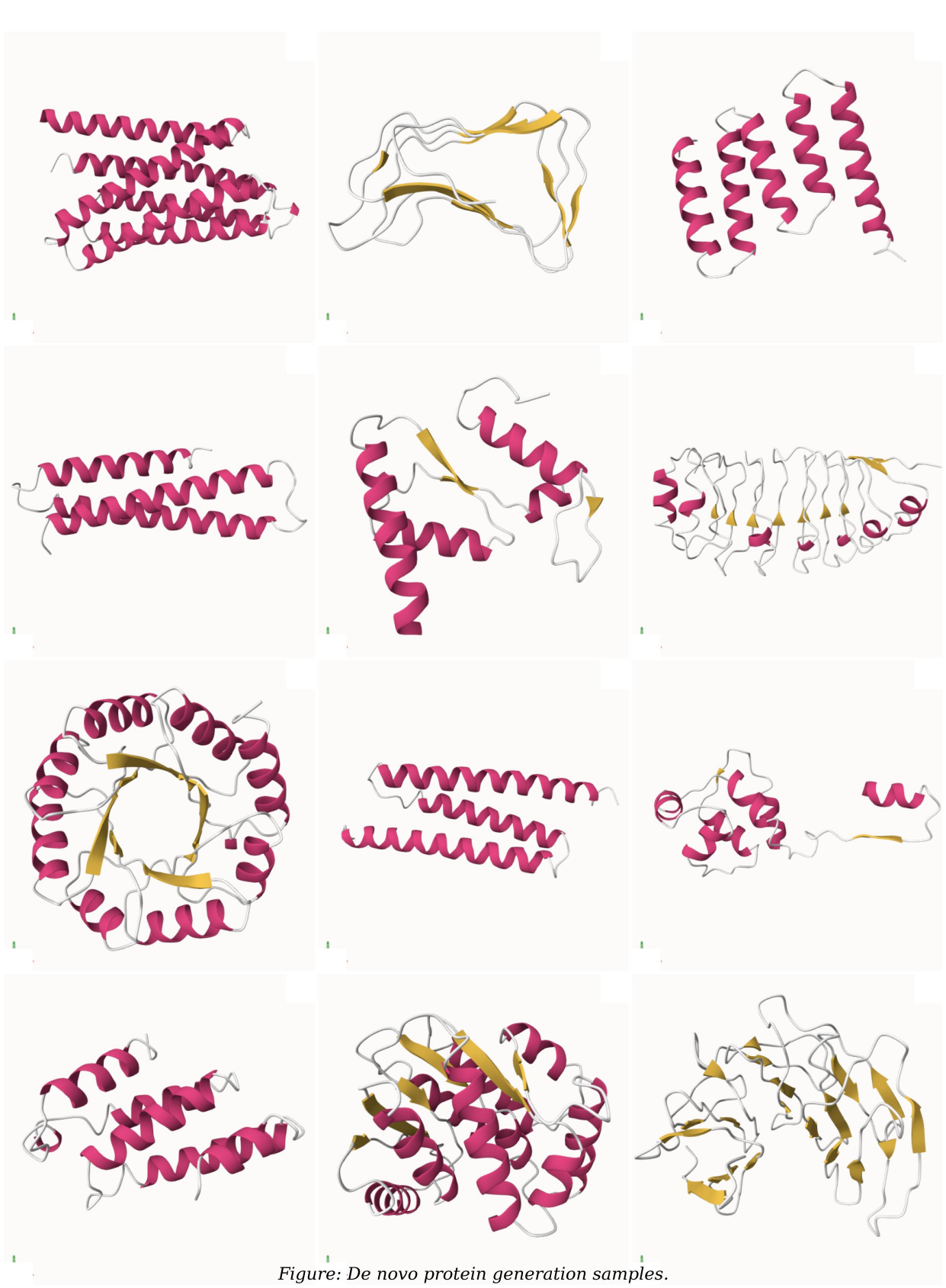}
    \caption{\textbf{Generated samples on AFDB-FS.} Unconditional
    protein backbones from the protein-only \ours prior; structures
    shown as cartoon traces of the C$\alpha$ chain.}
    \label{fig:samples_protein}
\end{figure}

Figures~\ref{fig:samples_qm9}--\ref{fig:samples_protein} show
unconditional samples drawn from the latent DiT prior of \ours across
all evaluated domains (QM9 small molecules, GEOM-Drugs drug-like
molecules, MP-20 crystals, and AFDB-FS proteins), decoded with the
corresponding flow-matching decoder. The figures provide illustrative samples; aggregate quality is
reported in the quantitative tables.

\clearpage

\clearpage

\clearpage

\end{document}